# Case-Based Subgoaling in Real-Time Heuristic Search for Video Game Pathfinding

**Vadim Bulitko** BULITKO@UALBERTA.CA
*Department of Computing Science, University of Alberta*
*Edmonton, Alberta, T6G 2E8, CANADA*
**Yngvi Björnsson** YNGVI@RU.IS
*School of Computer Science, Reykjavik University*
*Menntavegi 1, IS-101 Reykjavik, ICELAND*
**Ramon Lawrence** RAMON.LAWRENCE@UBC.CA
*Computer Science, University of British Columbia Okanagan*
*3333 University Way, Kelowna, British Columbia, V1V 1V7, CANADA*

## Abstract

Real-time heuristic search algorithms satisfy a constant bound on the amount of planning per action, independent of problem size. As a result, they scale up well as problems become larger. This property would make them well suited for video games where Artificial Intelligence controlled agents must react quickly to user commands and to other agents' actions. On the downside, real-time search algorithms employ learning methods that frequently lead to poor solution quality and cause the agent to appear irrational by re-visiting the same problem states repeatedly. The situation changed recently with a new algorithm, D LRTA*, which attempted to eliminate learning by automatically selecting subgoals. D LRTA* is well poised for video games, except it has a complex and memory-demanding pre-computation phase during which it builds a database of subgoals. In this paper, we propose a simpler and more memory-efficient way of pre-computing subgoals thereby eliminating the main obstacle to applying state-of-the-art real-time search methods in video games. The new algorithm solves a number of randomly chosen problems off-line, compresses the solutions into a series of subgoals and stores them in a database. When presented with a novel problem on-line, it queries the database for the most similar previously solved case and uses its subgoals to solve the problem. In the domain of pathfinding on four large video game maps, the new algorithm delivers solutions eight times better while using $57$ times less memory and requiring $14\%$ less pre-computation time.

## 1. Introduction

Heuristic search is a core area of Artificial Intelligence (AI) research and its algorithms have been widely used in planning, game-playing and agent control. In this paper we are interested in *real-time* heuristic search algorithms that satisfy a constant upper bound on the amount of planning per action, independent of problem size. This property is important in a number of applications including autonomous robots and agents in video games. A common problem in video games is searching for a path between two locations. In most games, agents are expected to act quickly in response to player's commands and other agents' actions. As a result, many game companies impose a constant time limit on the amount of path planning per move (e.g., one millisecond for *all* simultaneously moving agents).

While in practice this time limit can be satisfied by limiting problem size *a priori*, a scientifically more interesting approach is to impose a time per-move limit *regardless* of the problem size. Doing so severely limits the range of applicable heuristic search algorithms. For instance, static search algorithms such as A* (Hart, Nilsson, & Raphael, 1968), IDA* (Korf, 1985) and PRA* (Sturtevant & Buro, 2005; Sturtevant, 2007), re-planning algorithms such as D* (Stenz, 1995), anytime algorithms such as ARA* (Likhachev, Gordon, & Thrun, 2004) and anytime re-planning algorithms such as AD* (Likhachev, Ferguson, Gordon, Stentz, & Thrun, 2005) cannot guarantee a constant bound on planning time per action. This is because all of them produce a complete, possibly abstract, solution before the first action can be taken. As the problem increases in size, their planning time will inevitably increase, exceeding any *a priori* finite upper bound.

Real-time search addresses the problem in a fundamentally different way. Instead of computing a complete, possibly abstract, solution before the first action is taken, real-time search algorithms compute (or plan) only a few first actions for the agent to take. This is usually done by conducting a lookahead search of a fixed depth (also known as "search horizon", "search depth" or "lookahead depth") around the agent's current state and using a heuristic (i.e., an estimate of the remaining travel cost) to select the next few actions. The actions are then taken and the planning-execution cycle repeats (Korf, 1990). Since the goal state is not seen in most such local searches, the agent runs the risks of selecting suboptimal actions. To address this problem, real-time heuristic search algorithms update (or learn) their heuristic function over time.

The learning process has precluded real-time heuristic search agents from being widely deployed for pathfinding in video games. The problem is that such agents tend to "scrub" (i.e., repeatedly re-visit) the state space due to the need to fill in heuristic depressions (Ishida, 1992). As a result, solution quality can be quite low and, visually, the scrubbing behavior is perceived as irrational.

Since the seminal work on LRTA* (Korf, 1990), researchers have attempted to speed up the learning process. We briefly describe the efforts in the related work section. Here, we note that while the various approaches all brought about improvements, a breakthrough performance was achieved by virtually eliminating the learning process in D LRTA* (Bulitko, Luštrek, Schaeffer, Björnsson, & Sigmundarson, 2008). This was done by computing the heuristic with respect to a near-by subgoal as a distant goal. Offline, D LRTA* constructs a high-level graph of regions using state abstractions, calculates optimal paths between all region pairs, and then stores as subgoals the states where the paths cross region boundaries. During the online search, D LRTA* consults the database to find the next subgoal with respect to the current and goal regions. Since heuristic functions usually relax the problem (e.g., the Euclidean distance heuristic ignores obstacles on a map), they tend to be more accurate closer to a goal. As a result, a heuristic function with respect to a near-by goal tends to be more accurate and, therefore, requires less adjustment (i.e., learning). Consequently, the solution quality is improved and the scrubbing behavior is reduced.

In this paper, we adapt the idea of subgoaling and make the following four contributions. First, we simplify the pre-processing step of D LRTA*. Instead of using state abstraction to select subgoals, we employ a nearest-neighbour algorithm over a database of solved cases. Second, we introduce the idea of compressing a solution path into a series of subgoals so that each can be "easily" reached from the previous one. In doing so, we use hill-climbing as a proxy for the notion of "easy reachability by LRTA*". Third, we employ kd-trees in order to access the case base effectively. Finally, we evaluate the new algorithm empirically in large-scale problem spaces.

The new algorithm is called k Nearest Neighbor LRTA* (or kNN LRTA*) and, for the rest of the paper, we set $k = 1$. This paper extends our previous conference publication (Bulitko & Björnsson, 2009) in the following ways. We store multiple goals per path to reduce the number of database accesses and we use kd-trees to speed up each database access. Additionally, we make optimizations to the on-line component of kNN LRTA*: evaluating only a small number of most similar database entries, interrupting LRTA* if it starts learning excessively and engaging start and end path optimizations. On the empirical evaluation side, we now use native multi-million state static video game maps and compare our algorithm to a newly published state-of-the-art non-learning real-time search algorithm (Björnsson, Bulitko, & Sturtevant, 2009).

The rest of the paper is organized as follows. In Sections 2 and 3 we formulate the problem of real-time heuristic search and show how the core LRTA* algorithm can be extended with subgoal selection. Section 4 analyzes related research. Section 5 provides intuition for the new algorithm following with details and pseudocode in Section 6. In Section 7 we give theoretical analysis and, in Section 8, empirically evaluate the algorithm in the domain of pathfinding. Section 9 summarizes the empirical results. We then conclude with a discussion of current shortcomings and future work.

## 2. Problem Formulation

We define a heuristic search problem as a directed graph containing a finite set of states (vertices) and weighted edges, with a single state designated as the *goal state*. At every time step, a search agent has a single *current state*, a vertex in the search graph, and takes an *action* (or makes a *move*) by traversing an out-edge of the current state. By traversing an edge between states $s_1$ and $s_2$ the agent changes its current state from $s_1$ to $s_2$. We say that a state is *visited* by the agent if and only if it was the agent's current state at some point of time. As it is usual in the field of real-time heuristic search, we assume that path planning happens *between* the moves (i.e., the agent does not think while traversing an edge). The "plan a move" - "travel an edge" loop continues until the agent arrives at its goal state, thereby solving the problem.

Each edge has a positive cost associated with it. The total cost of edges traversed by an agent from its start state until it arrives at the goal state is called the *solution cost*. We require algorithms to be *complete* (i.e., produce a path from start to goal in a finite amount of time if such a path exists). In order to guarantee completeness for real-time heuristic search we make the assumption of safe explorability of our search problems. Specifically, all costs are finite and for any states $s_1, s_2, s_3$, if there is a path between $s_1$ and $s_2$ and there is a path between $s_1$ and $s_3$ then there is also a path between $s_2$ and $s_3$.

Formally, all algorithms discussed in this paper are applicable to any such heuristic search problem. To keep the presentation focused and intuitive as well as to afford a large-scale empirical evaluation, we will use a particular type of heuristic search problem, pathfinding in grid worlds, for the rest of the paper. We discuss applicability of the new methods we suggest to other heuristic search and general planning problems in Section 11.

In video-game map settings, states are vacant square grid cells. Each cell is connected to four cardinally (i.e., west, north, east, south) and four diagonally neighboring cells. Outbound edges of a vertex are moves available in the corresponding cell and in the rest of the paper we will use the terms *action* and *move* interchangeably. The edge costs are *defined* as 1 for cardinal moves and 1.4 for diagonal moves.[1]

An agent plans its next action by considering states in a local search space surrounding its current position. A *heuristic function* (or simply *heuristic*) estimates the (remaining) travel cost

1. We use 1.4 instead of the Euclidean $\sqrt{2}$ to avoid errors in floating point computations.

between a state and the goal. It is used by the agent to rank available actions and select the most promising one. In this paper we consider only *admissible* and *consistent* heuristic functions which do not overestimate the actual remaining cost to the goal and whose difference in values for any two states does not exceed the cost of an optimal path between these states. In this paper we use *octile distance* – the minimum cumulative edge cost between two vertices ignoring map obstacles – as our heuristic. This heuristic is admissible and consistent and uses $1$ and $1.4$ as the edge costs. An agent can modify its heuristic function in any state to avoid getting stuck in local minima of the heuristic function, as well as to improve its action selection with experience.

The defining property of real-time heuristic search is that the amount of planning the agent does per action has an upper bound that does not depend on the total number of states in the problem space. Fast planning is preferred as it guarantees the agent's quick reaction to a new goal specification. We measure mean *planning time* per action in terms of CPU time. We do not use the number of states expanded as a CPU-independent measure of time because the algorithms evaluated in this paper frequently perform time-consuming operations other than expanding states. Also note that while total planning time per problem is important for non-real-time search, it is irrelevant in video game pathfinding as we do not compute an entire path outright.

The second performance measure of our study is *sub-optimality* defined as the ratio of the solution cost found by the agent ($c$) to the minimum solution cost ($c^*$) minus one and times $100\%$: $\left(\frac{c}{c^*} - 1\right) \cdot 100$. To illustrate, suboptimality of $0\%$ indicates an optimal path and suboptimality of $50\%$ indicates a path $1.5$ times as costly as the optimal path.

## 3. LRTA*: The Core Algorithm

The core of most real-time heuristic search algorithms is an algorithm called Learning Real-Time A* (LRTA*) (Korf, 1990). It is shown in Figure 1 and operates as follows. As long as the goal state $s_{\text{global goal}}$ is not reached, the algorithm interleaves planning and execution in lines 4 through 7. In our generalized version we added a new step at line 3 for selecting goal $s_{\text{goal}}$ (the original algorithm uses $s_{\text{global goal}}$ at all times). We will describe the details of subgoal selection later in the paper. In line 4, a cost-limited breadth-first search with duplicate detection is used to find frontier states with cost up to $g_{\max}$ away from the current state $s$. For each frontier state $\hat{s}$, its value is the sum of the cost of a shortest path from $s$ to $\hat{s}$, denoted by $g(s, \hat{s})$, and the estimated cost of a shortest path from $\hat{s}$ to $s_{\text{goal}}$ (i.e., the heuristic value $h(\hat{s}, s_{\text{goal}})$). The state that minimizes the sum is identified as $s'$ in line 5. Ties are broken in favour of higher $g$ values. Remaining ties are broken in a fixed order. The heuristic value of the current state $s$ is updated in line 6 (we keep separate heuristic tables for the different goals and we never decrease heuristics). Finally, we take one step towards the most promising frontier state $s'$ in line 7.

LRTA* is a special case of value iteration or real-time dynamic programming (Barto, Bradtke, & Singh, 1995) and has a problem that has prevented its use in video game pathfinding. Specifically, it updates a single heuristic value per move on the basis of heuristic values of near-by states. This means that when the initial heuristic values are overly optimistic (i.e., too low), LRTA* will frequently re-visit these states multiple times, each time making updates of a small magnitude. This behavior is known as "scrubbing"[2] and appears highly irrational to an observer.

2. The term was coined by Nathan Sturtevant.

**LRTA***$(s_{\text{start}}, s_{\text{global goal}}, g_{\text{max}})$

$s \leftarrow s_{\text{start}}$
**while** $s \neq s_{\text{global goal}}$ **do**
    **if** no subgoal is selected or the current subgoal is reached **then** select a (new) subgoal $s_{\text{goal}}$
    generate successor states of $s$ up to $g_{\text{max}}$ cost, generating a frontier
    find a frontier state $s'$ with the lowest $g(s, s') + h(s', s_{\text{goal}})$
    update $h(s, s_{\text{goal}})$ to $g(s, s') + h(s', s_{\text{goal}})$
    change $s$ one step towards $s'$
**end while**

Figure 1: LRTA* algorithm with dynamic subgoal selection.

## 4. Related Research

Since the seminal work on LRTA* described in the previous section, researchers have attempted to speed up the learning process. Most of the resulting algorithms can be described by the following four attributes:

The **local search space** is the set of states whose heuristic values are accessed in the planning stage. The two common choices are full-width limited-depth lookahead (Korf, 1990; Shimbo & Ishida, 2003; Shue & Zamani, 1993; Shue, Li, & Zamani, 2001; Furcy & Koenig, 2000; Hernández & Meseguer, 2005a, 2005b; Sigmundarson & Björnsson, 2006; Rayner, Davison, Bulitko, Anderson, & Lu, 2007) and A*-shaped lookahead (Koenig, 2004; Koenig & Likhachev, 2006). Additional choices are decision-theoretic based shaping (Russell & Wefald, 1991) and dynamic lookahead depth-selection (Bulitko, 2004; Luštrek & Bulitko, 2006). Finally, searching in a smaller, abstracted state has been used as well (Bulitko, Sturtevant, Lu, & Yau, 2007).

The **local learning space** is the set of states whose heuristic values are updated. Common choices are: the current state only (Korf, 1990; Shimbo & Ishida, 2003; Shue & Zamani, 1993; Shue et al., 2001; Furcy & Koenig, 2000; Bulitko, 2004), all states within the local search space (Koenig, 2004; Koenig & Likhachev, 2006) and previously visited states and their neighbors (Hernández & Meseguer, 2005a, 2005b; Sigmundarson & Björnsson, 2006; Rayner et al., 2007).

A **learning rule** is used to update the heuristic values of the states in the learning space. The common choices are mini-min (Korf, 1990; Shue & Zamani, 1993; Shue et al., 2001; Hernández & Meseguer, 2005a, 2005b; Sigmundarson & Björnsson, 2006; Rayner et al., 2007), their weighted versions (Shimbo & Ishida, 2003), max of mins (Bulitko, 2004), modified Dijkstra's algorithm (Koenig, 2004), and updates with respect to the shortest path from the current state to the best-looking state on the frontier of the local search space (Koenig & Likhachev, 2006). Additionally, several algorithms learn more than one heuristic function (Russell & Wefald, 1991; Furcy & Koenig, 2000; Shimbo & Ishida, 2003).

The **control strategy** decides on the move following the planning and learning phases. Commonly used strategies include: the first move of an optimal path to the most promising frontier state (Korf, 1990; Furcy & Koenig, 2000; Hernández & Meseguer, 2005a, 2005b), the entire path (Bulitko, 2004), and backtracking moves (Shue & Zamani, 1993; Shue et al., 2001; Bulitko, 2004; Sigmundarson & Björnsson, 2006).

Given the multitude of proposed algorithms, unification efforts have been undertaken. In particular, Bulitko and Lee (2006) suggested a framework, called Learning Real Time Search (LRTS), to combine and extend LRTA* (Korf, 1990), weighted LRTA* (Shimbo & Ishida, 2003), SLA* (Shue & Zamani, 1993), SLA*T (Shue et al., 2001), and to a large extent, $\gamma$-Trap (Bulitko, 2004). In the dimensions described above, LRTS operates as follows. It uses a full-width fixed-depth local search space with transposition tables to prune duplicate states. LRTS uses a max of mins learning rule to update the heuristic value of the current state (its local learning space). The control strategy moves the agent to the most promising frontier state if the cumulative volume of heuristic function updates on a trial is under a user-specified quota or backtracks to its previous state otherwise.

While the approaches listed above all brought about various improvements, breakthrough performance came in the form of subgoaling. Since commonly used heuristics simplify the problem at hand (e.g., the octile distance in grid-world pathfinding ignores obstacles), using LRTA* with near-by subgoals effectively increases heuristic quality and thus reduces the amount of learning.

Although in general planning a goal is often represented as a conjunction of simple subgoals, to the best of our knowledge, the only real-time heuristic search algorithm to implement subgoaling is D LRTA* (Bulitko, Björnsson, Luštrek, Schaeffer, & Sigmundarson, 2007; Bulitko et al., 2008). In its pre-processing phase, D LRTA* uses the clique abstraction of Sturtevant and Buro (2005) to create a smaller search graph. The clique abstraction collapses a set of fully connected states into a single abstract state and can be applied iteratively to compute progressively smaller graphs. For example, a 2-level abstraction applies the clique abstraction to a graph that has already been abstracted once. Similarly, an $a$-level abstraction applies the clique abstraction $a$ times. If we assume that each abstraction reduces the graph by a constant factor $\alpha$, an $a$-level abstract graph would contain $\alpha^a$ times fewer states than the original graph. This abstraction technique in effect partitions the map into a number of regions, with each region corresponding to a single abstract state. Then for every pair of distinct abstract states, D LRTA* computes an optimal path between corresponding representative states (e.g., centroids of the regions) in the original non-abstracted space. The path is followed until it exits the region corresponding to the start abstract state. The entry state to the next region is recorded as the subgoal for the pair of abstract states. Once this pre-processing step is finished, D LRTA* runs LRTA* for a given problem but selects the subgoal recorded for the current and the goal regions. The off-line and on-line steps are illustrated in Figure 2.

The underlying intuition is that reaching the entry-to-the-next-region state requires LRTA* to navigate within a single region and is, therefore, "easy" with a default heuristic function. As a result, D LRTA* would rarely need to adjust its heuristic thereby virtually eliminating the costly learning process and the resulting scrubbing.

There are three key problems with D LRTA*. First, due to the fact that entry states (i.e., subgoals) have to be computed and stored for *each pair* of distinct regions, the number of regions has to be kept relatively small. In D LRTA* this is accomplished by applying the clique abstraction procedure multiple times so that the regions become progressively larger and fewer in number. A side effect is that regions will no longer be cliques and may, in fact, be quite complex in themselves. As a result, LRTA* may encounter heuristic depressions *within* a region (e.g., this would actually happen if LRTA* tries to go from $S$ to $E$ in the right diagram of Figure 2). Second, each state in the original space needs to be assigned to a region. Since the regions are irregular in shape, explicit membership records must be maintained. This may require as much additional memory as storing the original grid-based map. Third, clique abstraction is a non-trivial process and puts an extra programming burden on practitioners (e.g., game developers).

Another recent high-performance real-time search algorithm is Time-Bounded A* (TBA*) by Björnsson et al. (2009), a time-bounded variant of the classic A*. It expands states in an A* fashion using a closed list and an open list, away from the original start state, towards the goal until the goal state is expanded. However, unlike A* that plans a complete path before committing to the first action, TBA* time-slices the planning by interrupting its search periodically and acts. Initially, before a complete path to the goal is known, the agent takes an action that moves it towards the most promising state on the open list. If on a subsequent time slice an alternative most promising path is formed and the agent is not on that path, it backtracks its steps as necessary. This interleaving of planning, acting, and backtracking is done in such a way that both real-time behavior and completeness are ensured. The size of the time-slice is given as a parameter to the algorithm, using as a metric the number of states allowed to expand before the planning must be interrupted. Within a single time-slice, however, operations for both state expansions and backtracing the closed list (to form the path to the most promising state on the open list) must be performed. The cost of the latter type of operations is thus converted to state expansion equivalence (typically several backtracing steps can be performed at the same computational cost as a single state expansion). A key aspect of TBA* over LRTA*-based algorithms is that it retains closed and open lists over its planning steps. Thus, on each planning step it does not start planning from scratch, but continues with its open and closed lists from the previous planning step. Also, it does not need to update heuristics online to ensure completeness, nor does it require a precomputation phase. While the lack of precomputation is certainly its strong side, the negatives include high suboptimality if the amount of time per move is low and high on-line space complexity due to storing closed and open lists.

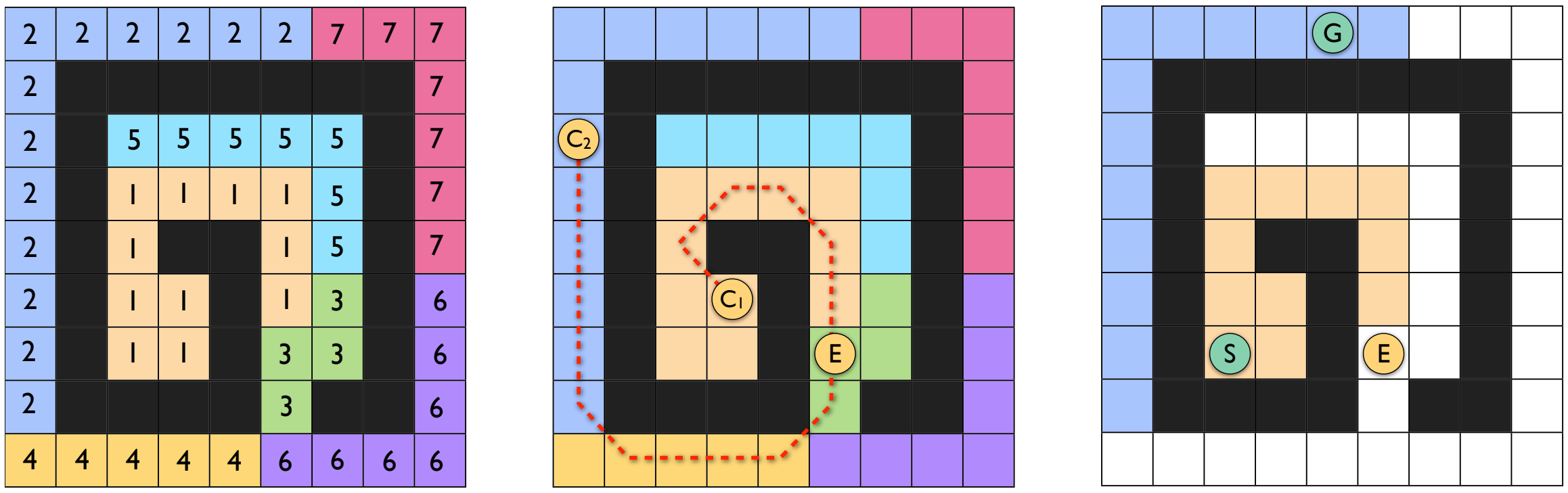

Figure 2: Example of D LRTA* operation. **Left:** off-line, the map is partitioned into seven regions (or abstract states). Each vacant cell is labeled with its region number. **Center:** off-line, an optimal path between centroids of two regions ($C_1$ and $C_2$) is computed and the entry state to the next region ($E$) is recorded as a subgoal for this pair of regions. **Right:** on-line, the agent intends to travel from $S$ to $G$, it determines the corresponding regions and sets the pre-computed entry state $E$ as its subgoal.

This research is related to work from the realm of non-real-time heuristic search where pattern databases are widely used to store pre-calculated distance information about abstractions of the original (ground) search space (Culberson & Schaeffer, 1998). A recent approach for using pre-calculated state-space information is to calculate true distances between selected state pairs and then use them whenever possible to make the distance estimates of the search guidance heuristic $h$ more informative. Two such enhanced heuristics are the *differential heuristic* (Cazenave, 2006; Sturtevant, Felner, Barrer, Schaeffer, & Burch, 2009) and the *canonical heuristic* (Sturtevant et al., 2009). In the former case, a true distance $d$ is pre-calculated from all states to a small subset of states $S$, so-called canonical states. During the on-line search the heuristic distance between any two arbitrary states $a$ and $b$ is calculated as the maximum of $h(a,b) = |d(a,s) - d(b,s)|$ over all canonical states $s \in S$. In the latter case, for each state in the state space the true distance $d$ to the closest canonical state is pre-calculated and stored and so is the true distance between all pairs of canonical states. During the search, the heuristic distance between any two states $a$ and $b$ is calculated as $h(a,b) = d(C(a), C(b)) - d(a, C(a)) - d(b, C(b))$ where $C(s)$ returns the closest canonical state to $s$. These heuristics may return a lower distance estimate than an unmodified heuristic, so in practice one chooses the maximum of the two. An idea similar to the canonical heuristic was proposed earlier in a more specialized context, where the heuristic function was improved by pre-calculating true distances between several strategically chosen passageways in a game map (Björnsson & Halldórsson, 2006). These heuristics are not used in real-time search.

There is a large volume of work on case-based planning (e.g., Nebel & Koehler, 1995). This includes path planning, where case-based approaches have been used to augment heuristic search for tasks such as route selection in road maps and mobile robot navigation. Such approaches typically pre-compute and store paths, as opposed to distances, between selected states, and then use them as model solutions for related pathfinding tasks in a case-based reasoning (CBR) fashion. One of the early works on combining search and case-based reasoning in pathfinding on road maps was done within the planning and learning system PRODIGY (Carbonell, Knoblock, & Minton, 1990), with the goal of generating near-optimal routes for an autonomous navigation vehicle trying to achieve multiple goals while driving in a city (Haigh & Veloso, 1993). The authors acknowledge the benefits of such an approach in the situation where it is necessary to interleave planning and execution. Subsequent work on case-based route selection has though mainly focused on augmenting non-interleaving path-planning algorithms, such as A* or Dijkstra, with the focus of the work on how best to build the case base, for example, how to identify, compute, and store paths to critical junctions that many paths pass through (Anwar & Yoshida, 2001; Weng, Wei, Qu, & Cai, 2009). As for mobile robot navigation, two heuristic search algorithms working in ground space and using a CBR-based approach were introduced by Branting and Aha (1995). The simpler one, when looking for a path between states $a$ and $b$, searches the pre-calculated case base for a path that contains both $a$ and $b$. If a match is found the best path is returned, otherwise a regular A* search is invoked to calculate the solution path. The second, and more elaborate, algorithm searches the case base for a match in the same fashion as the first, but if none is found, it adapts an existing case to fit the new task. This is done by using A* to join $a$ and $b$ to an existing path in the case base so that the new overall distance is minimized. There is still ongoing research in this area, for example, work on storing the case base as a graph structure called a case-graph that gradually builds a waypoint-like navigation network (Hodal & Dvorak, 2008). Note that these and many other existing algorithms are not real-time as they generate or modify complete plans.

## 5. Intuition for kNN LRTA*

In our design of kNN LRTA* we address the three shortcomings of D LRTA* listed earlier. In doing so, we identify two key aspects of a subgoal-based real-time heuristic search. First, we need to define a set of subgoals that would be efficient to compute and store off-line. Second, we need to define a way for the agent to find a subgoal relevant to its current problem on-line.

Intuitively, if an LRTA*-controlled agent is in the state $s$ going to the state $s_{\text{goal}}$ then the best subgoal is a state $s_{\text{ideal subgoal}}$ that resides on an optimal path between $s$ and $s_{\text{goal}}$ and can be reached by LRTA* along an optimal path with no state re-visitation. Given that there can be multiple optimal paths between two states, it is unclear how to computationally efficiently detect the LRTA* agent's deviation from an optimal path *immediately after it occurs*.

On the positive side, detecting state re-visitation can be done computationally efficiently by running a simple greedy hill-climbing agent. This is based on the fact that if a hill-climbing agent can reach a state $b$ from a state $a$ without encountering a local minimum or a plateau in the heuristic then an LRTA* agent can travel from $a$ to $b$ without state re-visitation (Theorem 5). Thus, we propose an efficiently computable approximation to $s_{\text{ideal subgoal}}$. Namely, we define the subgoal for a pair of states $s$ and $s_{\text{goal}}$ as the state $s_{\text{kNN LRTA* subgoal}}$ farthest along an optimal path between $s$ and $s_{\text{goal}}$ that can be reached by a simple hill-climbing agent (defined rigorously in the following section). In summary, we select subgoals to eliminate any scrubbing (Theorem 5) but do not guarantee that the LRTA* agent keeps on an optimal path between the subgoals (Theorem 6). In practice, however, only a tiny fraction of our subgoals are reached by the hill-climbing agent suboptimally and even then the suboptimality is minor.

This approximation to the ideal subgoal allows us to effectively compute a series of subgoals for a given pair of start and goal states. Intuitively, we *compress* an optimal path into a series of key states such that each of them can be reached from its predecessor without scrubbing. The compression allows us to save a large amount of memory without much impact on time-per-move. Indeed, hill-climbing from one of the key states to the next requires inspecting only the immediate neighbors of the current state and selecting one of them greedily. The re-visitation-free reachability of one subgoal from another addresses the first key shortcoming of D LRTA* where the agent may get "trapped" within a single complex region and thus be unable to reach its prescribed subgoal.

However, it is still infeasible to compute and then compress an optimal path between *every* two distinct states in the original search space. We solve this problem by compressing only a pre-determined fixed number of optimal paths between random states off-line. Then on-line kNN LRTA*, tasked with going from $s$ to $s_{\text{goal}}$, retrieves the most similar compressed path from its database and uses the associated subgoals. We define (dis-)similarity of a database path to the agent's current situation as the maximum of the heuristic distances between $s$ and the path's beginning and between $s_{\text{goal}}$ and the path's end. We use maximum because we would like *both* ends of the path to be heuristically close to the agent's current state and the goal respectively. Indeed, the heuristic distance ignores walls and thus a large heuristic distance to the path's either end tends to make that end hill-climbing unreachable.

Note that high similarity (i.e., both distances being low) does not guarantee that the path will be useful to the kNN LRTA* agent. For instance, the beginning of the path can be heuristically very close to the agent but on the other side of a long wall, making it unreachable without a lot of learning and the associated scrubbing. To address this problem we complement the fast-to-compute similarity metric with more computationally demanding move-limited reachability checks as detailed below.

We illustrate this intuition with a simple example. Figure 3 shows kNN LRTA* operation off-line. On this map, two random start and goal pairs are selected and optimal paths are computed between them. Then each path is compressed into a series of subgoals such that each of the subgoals can be reached from the previous one via hill-climbing. The path from $S_1$ to $G_1$ is compressed into two subgoals and the other path is compressed into a single subgoal.

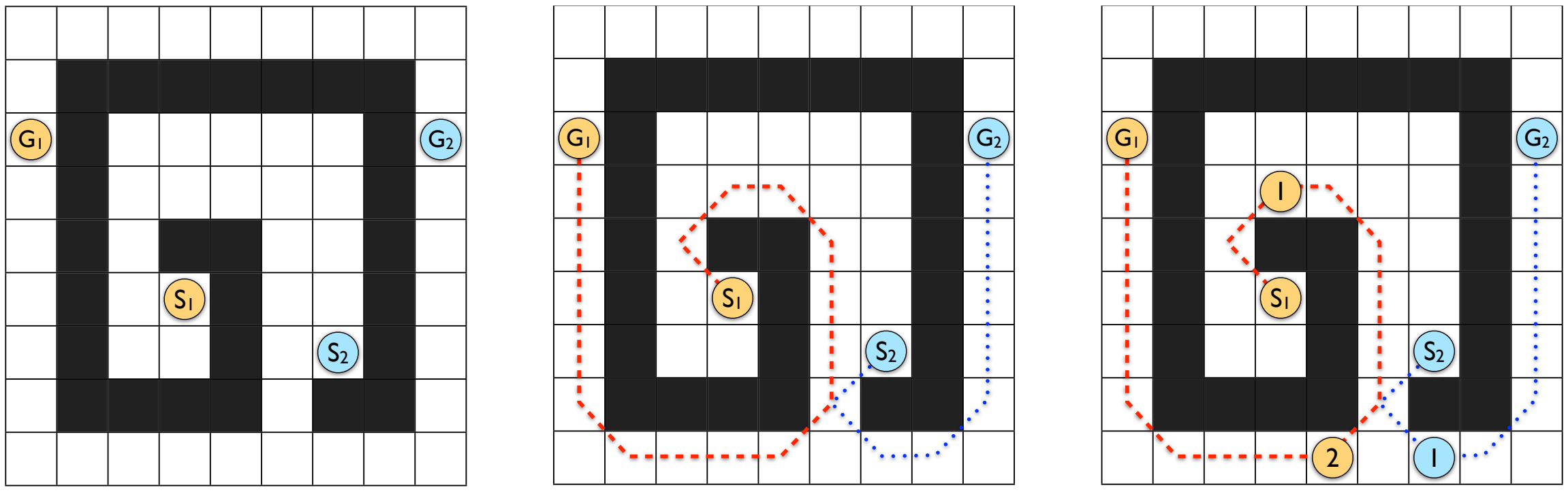

Figure 3: Example of kNN LRTA* off-line operation. **Left:** two subgoals (start,goal) pairs are chosen: $(S_1, G_1)$ and $(S_2, G_2)$. **Center:** optimal paths between then are computed by running A*. **Right:** the two paths are compressed into a total of three subgoals.

Once this database of two records is built, kNN LRTA* can be tasked with solving a problem on-line. In Figure 4 it is tasked with going from the state $S$ to the state $G$. The database is scanned and similarity between $(S, G)$ and each of the two database records is determined. The records are sorted by their similarity: $(S_1, G_1)$ followed by $(S_2, G_2)$. Then the agent runs reachability checks: from $S$ to $S_i$ and from $G_i$ to $G$ where $i$ runs the database indices in the order of record similarity. In this example, $S_1$ is found unreachable by hill-climbing from $S$ and thus the record $(S_1, G_1)$ is discarded. The second record passes hill-climbing checks and the agent is tasked with going to its first subgoal (shown as 1 in the figure).

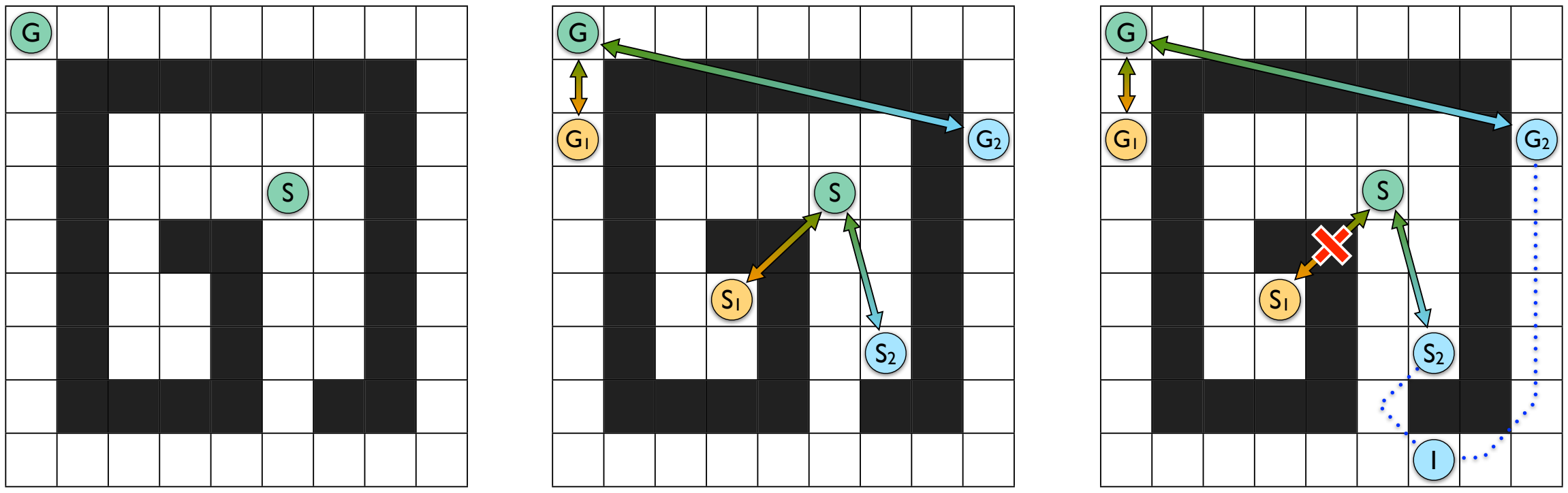

Figure 4: Example of kNN LRTA* on-line operation. **Left:** the agent intends to travel from $S$ to $G$. **Center:** similarity of $(S, G)$ to $(S_1, G_1)$ and $(S_2, G_2)$ is computed. **Right:** while $(S_1, G_1)$ is more similar to $(S, G)$ than $(S_2, G_2)$, its beginning $S_1$ is not reachable from $S$ via hill-climbing and hence the record $(S_2, G_2)$ is selected and the agent is tasked with going to subgoal 1.

The similarity plus hill-climbing check approach makes the state abstraction of D LRTA* unnecessary, thereby addressing its other two key shortcomings: high memory requirements and a complex pre-computation phase.

## 6. kNN LRTA* in Detail

In this section we flesh out kNN LRTA* in enough detail for other researchers to implement it. We start with a basic version and then describe several significant enhancements.

### 6.1 Basic kNN LRTA*

kNN LRTA* consists of two parts: database pre-computation (off-line) and LRTA* with dynamically selected subgoals (on-line). Pseudocode for the off-line part is presented in Figure 5. The top-level function **computeSubgoals** takes a user-controlled parameter $N$ and a search graph (e.g., a grid-based map in pathfinding) and builds a subgoal database of $N$ compressed paths. Each path is generated in line 4 from start and goal states randomly chosen in line 3. If the path does not exist or is too short (line 5), we discard it and re-generate the start and goal states. The compression takes place in the function **compress**, which returns a sequence of states $\Gamma_p = \left(s_{\text{start}}, s^1_{\text{subgoal}}, \dots, s^{n_p}_{\text{subgoal}}, s_{\text{goal}}\right)$ where $n_p \geq 0$ is the number of subgoals (line 6). The sequence $\Gamma_p$ is the compressed representation of path $p$ and forms a single record in the subgoal database (line 7).

```
subgoal database ← computeSubgoals(N, G)
1  subgoal database ← ∅
2  for n = 1, ..., N do
3      generate a random pair of states (s_start, s_goal)
4      compute an optimal path p from s_start to s_goal with A*
5      if p = ∅ ∨ |p| < 3 then go to step 3 end if
6      Γ_p ← compress(p)
7      add Γ_p to the subgoal database
8  end for
```

Figure 5: kNN LRTA* off-line: building a subgoal database.

Pseudocode for the function **compress** is found in Figure 6. It takes the path $p = (s_{\text{start}}, \dots, s_{\text{goal}}) = (s_1, \dots, s_t)$ as an argument and returns a subset $\Gamma$ of it — the states reachable from each other via hill-climbing (and thus without scrubbing). The code builds the sequence $\gamma$ of indices of states which will then be put into $\Gamma$ as subgoals. As long as the path is not exhausted (line 2), the next candidate subgoal is defined by the index $i$ in line 3. Note that the state with the index $i = \textbf{end}(\gamma)+1$ is always hill-climbing reachable from the state with the index $\textbf{end}(\gamma)$ because these two states are immediate neighbours. We then run a binary search defined by the scope of indices $[l, r]$ in lines 4 and 5. The middle of the scope is calculated in line 7 and its hill-climbing reachability from the latest computed subgoal $s_{\textbf{end}(\gamma)}$ is checked in line 8. If the middle is indeed hill-climbing reachable then the scope is moved to the upper half (line 10) and the candidate subgoal is updated (line 9). Otherwise, the scope of the binary search is moved to the lower half in line 12. Once the binary search is completed, the candidate subgoal is added to $\gamma$ in line 15.[3] We convert the sequence of indices $\gamma$ into the sequence of states $\Gamma$ in line 17.

The function **reachable**($s_a, s_b$) checks if a hill-climbing agent can reach the state $s_b$ from the state $s_a$. The pseudocode is found in Figure 7. We start climbing from the state $s_a$ (line 1). As long

3. We use parentheses with set operations to indicate that $\gamma$ is an *ordered* set.

$\Gamma \leftarrow$ **compress**$((s_1, \ldots, s_t))$

$\gamma \leftarrow (1)$
**while** $t \notin \gamma$ **do**
  $i \leftarrow$ **end**$(\gamma) + 1$
  $l \leftarrow i + 1$
  $r \leftarrow t$
  **while** $l \leq r$ **do**
    $m \leftarrow \lfloor \frac{l+r}{2} \rfloor$
    **if reachable**$(s_{\textbf{end}(\gamma)}, s_m)$ **then**
      $i \leftarrow m$
      $l \leftarrow m + 1$
    **else**
      $r \leftarrow m - 1$
    **end if**
  **end while**
  $\gamma \leftarrow \gamma \cup (i)$
**end while**
$\Gamma \leftarrow s_\gamma$

Figure 6: kNN LRTA* off-line: compressing a path into a sequence of subgoals.

$\rho \leftarrow$ **reachable**$(s_a, s_b)$

1 $s \leftarrow s_a$
2 **while** $s \neq s_b$ **do**
3   generate immediate successor states of $s$, generating a frontier
4   **if** $h(s) \leq \min_{s'' \in \text{frontier}}(h(s''))$ **then break**
5   find the frontier state $s'$ with the lowest $g(s, s') + h(s', s_b)$
6   $s \leftarrow s'$
7 **end while**
8 $\rho \leftarrow (s = s_b)$

Figure 7: Checking if one state is reachable from another. When this function is called on-line, a fixed cap is put on the number of iterations in the **while** loop.

as the goal is not reached (line 2), we generate immediate successors of the current state (line 3) and check if we are in a local heuristic minimum or a plateau (line 4). If so we terminate our climb and declare that $s_b$ is not hill-climbing reachable from $s_a$. Otherwise we climb towards a frontier state with the lowest $g + h$ value (lines 5 and 6). We use $g + h$ instead of $h$ to make the move selection correspond to that of LRTA*. Additionally, the ties are broken in exactly the same way as they are with the LRTA* algorithm in Figure 1. Note that whenever the function **reachable** is called in the on-line phase, we impose a fixed cutoff on the number of steps hill-climbing is allowed to travel. This is done to place an upper bound on the time complexity of the reachability check independent of the number of states in the search graph, as required by real-time operation.

In the on-line phase of kNN LRTA*, we run LRTA* as per Figure 1. Dynamic subgoal selection (line 3) is done as per pseudocode in Figure 8. Given a start and goal state, we scan our subgoal database and, for each record, compute heuristic distance between our start state and the record's first state as well as heuristic distance between our goal state and the record's last state. As we mentioned earlier, we define the (dis-)similarity between our problem and the record as the maximum of the two heuristic distances. This is done so that similar records are such where the start and end are *both* close to the agent's current position and its goal in terms of the heuristic distance.

All database records are sorted by their similarity to the agent's current and global goal states (line 1) and, starting with the most similar record, we check if its start and end are hill-climbing reachable from the agent's current state and the agent's global goal respectively (line 4). If either reachability check fails, we go onto the next record. Otherwise, we stop the database search (line 6). If we exhaust the database and find no reachable record, we resort to the global goal (line 9). Once a record is found, all its subgoals are fed one by one to LRTA* in line 3 of Figure 1.

The intuition is that our similarity metric uses heuristic distance and, therefore, ignores some constraints of the problem (e.g., walls in grid-based pathfinding). Thus, a database record with a high similarity value may not be relevant to the agent's situation as its start and goal may be on the other side of a wall which means that its subgoals will not be reachable by LRTA* without scrubbing and therefore are useless to the agent.

$r \leftarrow$ **selectSubgoals**$(s, s_{\text{global goal}})$

1 $(r_1, \dots, r_N) \leftarrow$ database records from most to least similar<br>
2 **for** $i = 1, \dots, N$ **do**<br>
3     retrieve $r_i = (s_{\text{start}}, \dots, s_{\text{end}})$<br>
4     **if reachable**$(s, s_{\text{start}})$ and **reachable**$(s_{\text{end}}, s_{\text{global goal}})$ **then**<br>
5         $r \leftarrow r_i$<br>
6         **return**<br>
7     **end if**<br>
8 **end for**<br>
9 $r \leftarrow \left(s, s_{\text{global goal}}\right)$

Figure 8: kNN LRTA* on-line: selecting subgoals.

### 6.2 Enhanced kNN LRTA*

We have presented the basic kNN LRTA* algorithm. In this section we introduce six enhancements.

First, before selecting a database record in the function **selectSubgoals**, we check if the global goal is reachable from the agent's current state. This is done by calling the function **reachable**. If the global goal is indeed reachable via move-limited hill-climbing then we set it as the agent's goal and do not look for a subgoal. Otherwise, we turn to the database for subgoals.

Second, having selected a database record in the routine **selectSubgoals**, we run a reachability check between the agent's current state and the first subgoal in the record. If the first subgoal is reachable then we set it as the goal for LRTA*. Otherwise, we set the LRTA* to go to the start state of the record which is already checked to be reachable within the function **selectSubgoals**.

Third, when LRTA* reaches the last subgoal (i.e., the state in the record immediately prior to the end of the record), it checks if the global goal is reachable from it. If so, the global goal is used as the next subgoal. Otherwise, the agent heads for the end of the record from which it can reach the global goal as guaranteed by the record selection criteria.

Each of the first three enhancements addresses a trade-off between path optimality and planning time per move. Specifically, calling the function **reachable**, while real-time, increases kNN LRTA* planning time per move but, at the same time, leads to a potentially shorter solution due to better subgoal selection. Recall that the function **reachable** satisfies the real-time operation constraint because we place an *a priori* limit on the number of moves it can take.

Reachability checks constitute a substantial portion of kNN LRTA*'s planning time per move. The other substantial contributor is accessing the record database and computing record similarity. The basic algorithm described above always computes the similarity for all database records and, in the worst case, runs reachability checks for all records in the function **selectSubgoals**. While this does not depend on the search graph size and is thus real-time, we can still speed it up as follows.

The fourth enhancement is to run reachability checks only for a fixed number of most similar records. This can be done simply by substituting the total number of database records $N$ with a fixed constant $M \leq N$ in line 2 of Figure 8. The intuition is that only fairly similar records are worth checking for reachability.

When $M \ll N$ this enhancement can substantially reduce the amount of planning time taken up by reachability checks. However, the similarity is still computed for all records in the database (line 1 in Figure 8). The fifth enhancement speeds this step up by employing kd-trees instead of a linear database scan. A kd-tree (Moore, 1991) is a spatial tree index that can have a sublinear time complexity for nearest-neighbor searches. Specifically, our kd-tree indexes start and end states of the subgoal database records. Each tree node is thus a four-tuple $(x_{\text{start}}, y_{\text{start}}, x_{\text{end}}, y_{\text{end}})$. The index works by dividing the search space along a dimension at each level of the tree. The search space is divided on $x_{\text{start}}$ below the root node of the tree, $y_{\text{start}}$ at the next level down, $x_{\text{end}}$ on the next level, $y_{\text{end}}$ on the next, and then the cycle repeats. For example, if the root node is $(4, 5, 8, 9)$, then the start state has coordinates $(4, 5)$ and the end state has coordinates $(8, 9)$. Further, any nodes in the left subtree have $x_{\text{start}} \leq 4$, and nodes in the right subtree have $x_{\text{start}} > 4$.

To illustrate, consider the tree in Figure 9 and a subgoal record whose start state is $(8, 4)$ and whose goal state is $(4, 9)$. This records will be represented by a kd-tree node $(8, 4, 4, 9)$. It is in the right subtree of the root as its $x_{\text{start}} = 8$ which is greater than the root's value of $4$. It is in the left subtree of the next node as its value of $4$ for $y_{\text{start}}$ is less than the node's value of $5$. At the third level, it is in the left subtree as its value of $4$ for $x_{\text{end}}$ is less than 6. Finally, it is the right subtree of its parent at level four because its value of $y_{\text{end}} = 9$ is greater than the 8 of its parent.

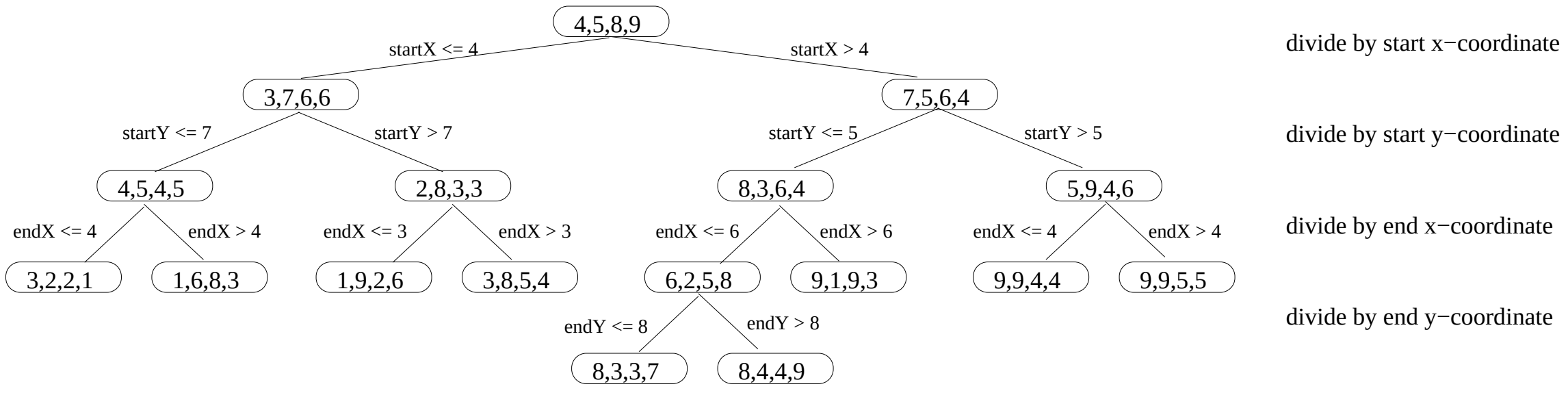

Figure 9: A kd-tree for database access.

This structure allows nearest-neighbors to be computed without searching all paths in the tree index by eliminating some subtrees based on distance. For instance, if the search currently has the best $M$ records found so far, and it encounters a node in the tree where it can be guaranteed that all nodes are farther away than those $M$ records from the search target, then that subtree is not searched. The nearest-neighbor search algorithm is explained by Moore (1991). Note that the kd-tree index works for regular grid pathfinding problems but not necessarily all heuristic search problems. For instance, high-cost edges connecting states with similar coordinates or low-cost edges connecting states with distant coordinates would present a problem for the kd-tree index.

Given a subgoal database, we build a kd-tree to index it off-line and store it together with the database. On-line, we use the kd-tree to identify the $M$ records relevant to the agent's current start and goal states (line 1 in Figure 8). We then compute the similarity metric only for these $M$ records.

The sixth enhancement deals with the case where kNN LRTA* is unable to find a subgoal and resorts to its global goal. This happens in the function **selectSubgoals** (line 9 of Figure 8). A failure to find a subgoal is caused by none of the $M$ most similar records passing our reachability checks. Having to resort to a global goal indicates an insufficient database coverage of the current area in the space of start and goal state pairs. Given that records are compressed optimal paths between randomly generated start and goal states, database coverage is likely to be uneven. Thus resorting to a global goal should not be a permanent step as the agent traveling to a global goal is likely to enter an area covered by the database sooner or later. At that point, the record selection process can be repeated, hopefully resulting in a database hit. We implement this intuition in kNN LRTA* by imposing a travel quota on LRTA* after the function **selectSubgoals** fails to find a reachable record. The quota is computed as a heuristic distance between the agent's current state and the global goal multiplied by a fixed constant greater than 1. Once the agent exhausts its quota, **selectSubgoals** is called again. If it fails to find a reachable subgoal for a second time in a row, the quota is set to infinity leading to no further interruptions. This is necessary to guarantee completeness. Additionally, interrupting LRTA* indefinitely many times increases average planning time per move due to subgoal selection attempts.

We have also experimented with the idea that a database record of the form $(s_1, \ldots, s_n)$ does not have to be used in its entirety. Indeed, any of its fragments (i.e., $(s_i, \ldots, s_j)$ where $1 \leq i < j \leq n$) can be used within kNN LRTA* in the same fashion as the entire record. We implemented this idea by running the kd-tree search over all fragments of database records in addition to the whole records. The results were disappointing in several ways. First, the kd-tree algorithm becomes more complex and the kd-tree query time increases. Second, record fragments "crowd" the $M$ hits that the kd-tree returns and for which the similarity metric is computed. In practice this means that the kd-tree returns $M$ similar but not hill-climbing unreachable records and, thus, causes kNN LRTA* to resort to its global goal more often. This can be fixed by increasing $M$ accordingly but then the similarity computation and the hill-climbing checks become more costly.

## 7. Theoretical Analysis

In this section we prove completeness of the algorithm and analyze its complexity.

### 7.1 Off-line Complexity

Off-line kNN LRTA* generates $N$ records in a space of $S$ states. Let the diameter of the space (i.e., the number of states along the longest possible shortest path between two states) be $\delta_S$.

**Theorem 1** Off-line worst-case space complexity of kNN LRTA* is $O(N\delta_S + S)$.

**Proof.** In the worst case each optimal path kNN LRTA* generates between randomly selected start and goal is $\delta_S$ long and is minimally compressible. Minimum compression means that every state on the path is stored. If all $N$ records have this property then the total amount of database storage is $O(N\delta_S)$. Additionally, A* is run for each record and has the worst-case space complexity of $S$. □

**Theorem 2** Off-line worst-case time complexity of kNN LRTA* is $O(NS \log S + N \log N)$.

**Proof.** kNN LRTA* runs A* to compute an optimal path for $N$ pairs of randomly generated start and goal states. With a consistent heuristic and other constraints from our problem formulation, A*'s worst case time complexity is $O(S \log S)$. Since $\delta_S \leq S$, A*'s complexity dominates the worst-case time complexity of the function **compress** which is $O(\delta_S \log \delta_S)$. Additionally, building a kd-tree takes $O(N \log N)$. □

### 7.2 On-line Complexity

In this section we assume that LRTA* generates all immediate neighbors of its current state and only them on each move. In our grid pathfinding this can be easily accomplished by setting $g_{\max} = 1.4$. More generally, this can be guaranteed by substituting line 4 in Figure 1 with "generate immediate successor states of $s$".

**Theorem 3** kNN LRTA*'s on-line worst-case space complexity is $O(d_{\max} + S)$ where $d_{\max}$ is the maximum out-degree of any vertex and $S$ is the total number of states.

**Proof.** The open list of kNN LRTA* is at most the maximum number of immediate neighbors of any state (i.e., $d_{\max}$). As LRTA* learns, it has to store updated heuristic values, of which there are no more than $S$. Hence the overall space complexity is $O(d_{\max} + S)$. Note that in grid pathfinding $d_{\max} \ll S$ and $d_{\max}$ does not increase with map size, thereby reducing the upper bound to $O(S)$. □

**Theorem 4** kNN LRTA*'s per-move worst-case time complexity is $O(d_{\max} + N + M \log M)$ where $d_{\max}$ is the maximum out-degree of any vertex, $N$ is the total number of records in the database and $M$ is the number of candidate records selected by the kd-tree.

**Proof.** On each move kNN LRTA* invokes LRTA* which generates at most $d_{\max}$ states. On some moves, kNN LRTA* additionally searches its database to find the appropriate record. The database search starts with querying the kd-tree for $M$ records ($M \leq N$). While balanced kd-trees can have time complexity sub-linear in $N$, the worst case time for this step is still $O(N)$. We then sort the $M$ records by their similarity in $O(M \log M)$ time. Finally, move-limited hill-climbing checks are run for the records, collectively taking no more than $O(M)$ time. Thus, the overall per-move time complexity is $O(d_{\max} + N + M \log M)$ in the worst case. □

Note that this bound does not depend on $S$ and, therefore, makes kNN LRTA* real-time by our definition. Also note that in grid pathfinding $d_{\max} \ll N$ and $M \ll N$ which makes kNN LRTA*'s per move time complexity simply $O(N)$.

### 7.3 Completeness

**Theorem 5** For any two states $s_1$ and $s_2$, if $s_2$ is hill-climbing reachable from $s_1$ then an LRTA* agent starting in $s_1$ will reach $s_2$ without state re-visitation (i.e., scrubbing).

**Proof.** First, we show that if the hill-climbing agent (as specified in the function **reachable** in Figure 7) can reach $s_2$ from $s_1$ then it can never re-visit any states on its way. Suppose, that is not the case. Then there exists a state $s_3$ re-visited by the hill-climbing agent. Because our ties are broken in a fixed order, once the hill-climbing agent arrives at $s_3$ for the second time, it will continue following the same path as it did after the first visit and will, therefore, arrive at $s_3$ for the third time and so on. In other words, it will be in an infinite loop re-visiting $s_3$ repeatedly. This contradicts the fact that it was able to reach $s_2$.

From here we conclude that the path between $s_1$ and $s_2$ followed by the hill-climbing agent is free of repeated states. We now have to show that the LRTA* agent starting in $s_1$ will follow exactly the same path as the hill-climbing agent. Observe that the only difference between our hill-climbing agent (Figure 7) and the LRTA* agent (Figure 1) is the heuristic update rule in line 6 of the latter figure. The update rule can only increase heuristic values (i.e., make them less attractive to the agent) and only in already visited states. Since the hill-climbing agent never re-visits its states while traveling between $s_1$ and $s_2$, any increase in the heuristic values caused by LRTA* does not affect LRTA*'s move choice (line 5 in Figure 1). As a result, LRTA* will follow precisely the same path between $s_1$ and $s_2$ as the hill-climbing agent and thus will not re-visit any states. □

**Theorem 6** There exist two states $s_1$ and $s_2$ such that $s_2$ is hill-climbing reachable from $s_1$ but the path that the hill-climbing agent follows is not optimal (i.e., shortest).

**Proof.** The proof is constructive and presented in Figure 10. The darkened cells are walls. The hill-climbing agent, starting in the state $s_1$ will "hug" the wall on its way to the state $s_2$.The resulting path has the cost of $16.4$. The optimal path, however, takes advantage of diagonal moves by making a non-greedy move and going around the wall above the agent. Its cost is $15$. □

**Theorem 7** kNN LRTA* is complete for any size of its subgoal database if the underlying kNN LRTA* generates at least all immediate neighbors of the current state.

**Proof.** To prove completeness we need to show that for any pair of states $s_1$ and $s_2$, if there is a path between $s_1$ and $s_2$, kNN LRTA* will reach $s_2$ from $s_1$.

Given a problem, the subgoal selection module of kNN LRTA* (Figure 8) will either return a record of the form $r = (s_{\text{start}}, \dots, s_{\text{end}})$ or instruct LRTA* to go to the global goal. In the latter case, kNN LRTA* is complete because the underlying LRTA* is complete (Korf, 1990) as long as it generates all immediate neighbors of its current state.

In the former case, LRTA* is guaranteed to reach either $s_{\text{start}}$ or the first subgoal of $r$ due to the way $r$ is selected. Once any of the states in $r$ is reached, LRTA* is guaranteed to reach the subsequent states due to the completeness of the basic LRTA* and the way the subgoals are generated. Note that the interruptibility enhancement does not interfere with completeness because we can interrupt going for a global goal at most once. □

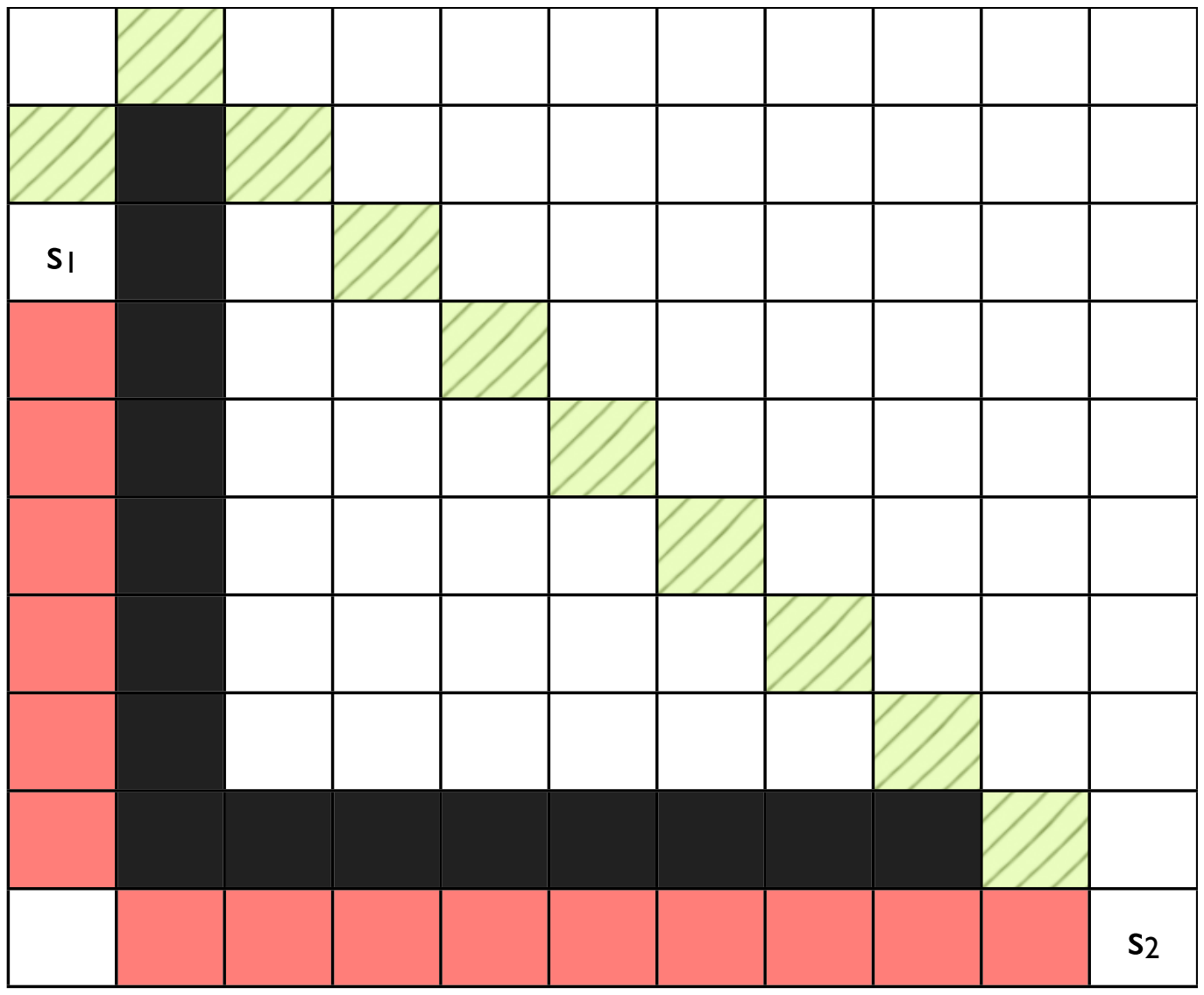

Figure 10: Hill-climbing reachability does not guarantee optimality.

## 8. Empirical Evaluation

Pathfinding in video games is a challenging task, frequently requiring many units to plan their paths simultaneously and to react promptly to user commands. The task is made even more challenging by ever-growing map sizes and little computational resources allocated to in-game AI. Accordingly, most recent work in the field of real-time heuristic search uses video game pathfinding as a testbed.

### 8.1 Test Problems

Maps modelled after game levels from *Baldur's Gate* (BioWare Corp., 1998) and *WarCraft III: Reign of Chaos* (Blizzard Entertainment, 2002) have been a common choice (e.g., Sturtevant & Buro, 2005; Bulitko et al., 2008). These maps, however, are small by today's standards and do not represent the state of the industry. For this paper, we developed a new set of maps modelled after game levels from *Counter-Strike: Source* (Valve Corporation, 2004), a popular on-line first-person shooter. In this game the level geometry is specified in a vector format. We developed software to convert it to a grid of an arbitrary resolution. While previous papers commonly used maps in the range of $10^4$ to $10^5$ grid cells (e.g., between $150 \times 141$ and $512 \times 512$ cells in Sturtevant, 2007; Bulitko & Björnsson, 2009), our new maps have between nine and thirteen million vacant cells (i.e., states). This is a two to three orders of magnitude increase in size. As a point of reference, the entire road network of Western Europe used for state-of-the-art route planning has approximately eighteen million vertices (Geisberger, Sanders, Schultes, & Delling, 2008).

The experiments in this paper were run on a set of $1000$ randomly generated problems across the four maps shown in Figure $11$. There were $250$ problems on each map and they were constrained to have solution cost of at least $1000$. The grid dimensions varied between $4096 \times 4604$ and $7261 \times 4096$ cells. For each problem we computed an optimal solution cost by running A*. The optimal cost was in the range of $[1003.8, 2999.8]$ with a mean of $1881.76$, a median of $1855.2$ and a standard deviation of $549.74$. We also measured the A* difficulty defined as the ratio of the number of states expanded by A* to the number of edges in the resulting optimal path. For the $1000$ problems, the

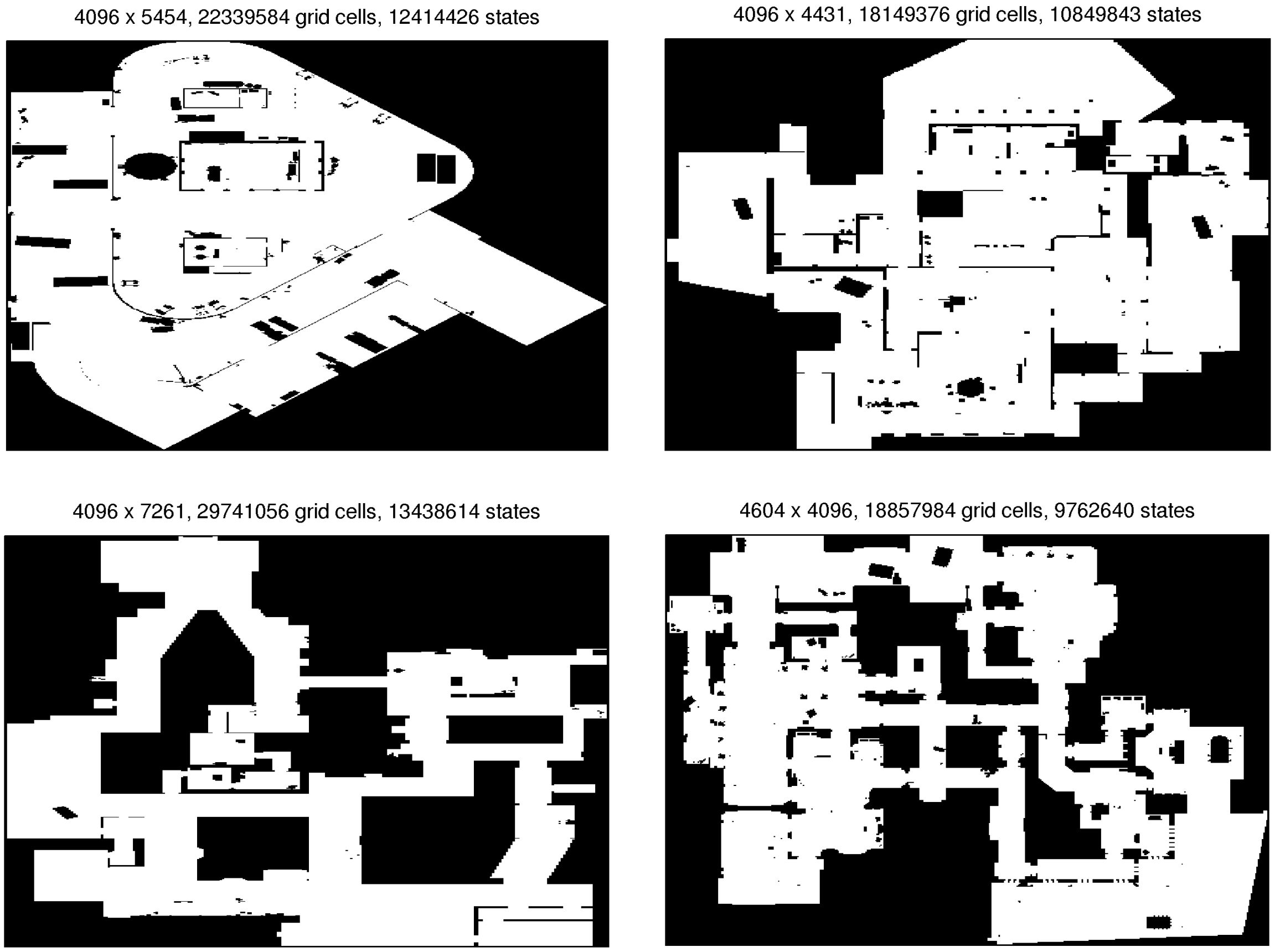

Figure 11: The maps used in our empirical evaluation.

A* difficulty was in the range of $[1, 199.8]$ with a mean of $62.60$, a median of $36.47$ and a standard deviation of $64.14$.

All algorithms compared were implemented in Java using common data structures as much as possible. We used Java version 6 under SUSE Enterprise Linux 10 on a 2.1 GHz AMD Opteron processor with 32 Gbytes of RAM. All timings are reported for single-threaded computations.

## 8.2 Algorithms Evaluated

We evaluated kNN LRTA* with the following parameters. Database size values were in $\{1000, 5000, 10000, 40000, 60000, 80000\}$ records. On-line, we allowed our hill-climbing test to climb for up to $250$ steps before concluding that the destination state is not hill-climbing reachable. This value was picked after some experimentation and had to be appropriate for the record density on the map. Indeed, a larger database requires fewer hill-climbing steps to maintain the likelihood of finding a hill-climbing reachable record for a given problem.

We ran reachability checks on the 10 most similar records.[4] Whenever **selectSubgoals** failed to find a matching record, we allowed LRTA* to travel towards its global goal up to 3 times the heuristic estimate of the remaining path. After that, LRTA* was interrupted and the second attempt to find an appropriate subgoal was run. LRTA*'s parameter $g_{\text{max}}$ was set to the cost of the most expensive edge (i.e., $1.4$) so that LRTA* generated only all immediate neighbors of its current state.

We also ran two recent high-performance real-time search algorithms to compare kNN LRTA* against: D LRTA* and TBA*. D LRTA* was run with the databases computed for abstraction levels of $\{9, 10, 11, 12\}$. TBA* was run with the time slices of $\{5, 10, 20, 50, 100, 500, 1000, 2000, 5000\}$. The cost ratio of expanding a state to backtracing was set to 10.

We chose the space of control parameters via trial and error, with three considerations in mind. First, we had to cover enough of the space to clearly determine the relationship between control parameters and algorithm's performance. Second, we attempted to establish the *pareto-optimal frontier* (i.e., determine which algorithms *dominate* others by simultaneously outperforming them along two performance measures such as time per move and suboptimality). Third, parameter values had to be such that we could run the algorithms in a practical amount of time (e.g., building a database for D LRTA*(8) would have taken us over 800 hours which is not practical). We detail our observations with respect to all three considerations below.

### 8.3 Solution Suboptimality and Per-Move Planning Time

We begin the comparisons by looking at average solution suboptimality versus average time per move. The left plot in Figure 12 shows the overall picture by plotting all algorithms and parameters. The right plot zooms in on a high-performance area. Table 1 shows the individual values. kNN LRTA* produces the highest quality solutions, followed by TBA*.

D LRTA* with its mean suboptimality of $819.72\%$ delivers paths which are about 9 times costlier than optimal paths. Such suboptimality is impractical in pathfinding and we included D LRTA*(9) in the right subplot of Figure 12 only to illustrate the substantial gap in solution quality between D LRTA* and kNN LRTA*. Optimality of D LRTA* solutions can be improved by lowering the abstraction level but the database pre-computation increases rapidly as we discuss below.

TBA* produces solutions substantially less costly than D LRTA* but cannot reach kNN LRTA* with the database size of 60 and 80 thousand records. Additionally, TBA* is noticeably slower per move as it expands more than one state and allocates some time to backtracking as well. The time per move can be decreased by lowering the value of cutoff but already with the cutoff of 10, TBA* produces unacceptably suboptimal solutions ($666.5\%$ suboptimal). As a result, kNN LRTA* dominates TBA* by outperforming it with respect to both measures. This is intuitive as TBA* does not benefit from subgoal precomputation.

On the other hand, D LRTA* stands non-dominated due to its low time per move. This is also intuitive as it does not have to scan the database for most similar records and then check hill-climbing reachability for them. The differences between D LRTA* and kNN LRTA* are, however, below 4 microseconds per move.

For the sake of reference, we also included A* results in the table. A* is not a real-time algorithm and its average time per move tends to increase with the number of states in the map. Also,

4. We have also experimented with querying the kd-tree for 100 most similar records and found a very minor improvement of suboptimality together with a significant increase in the mean time per move. This is because frequently a hill-climbing-reachable database record will be among the top 10 candidates and thus the extra time spent querying the kd-tree for 90 more records and then sorting them is wasted.

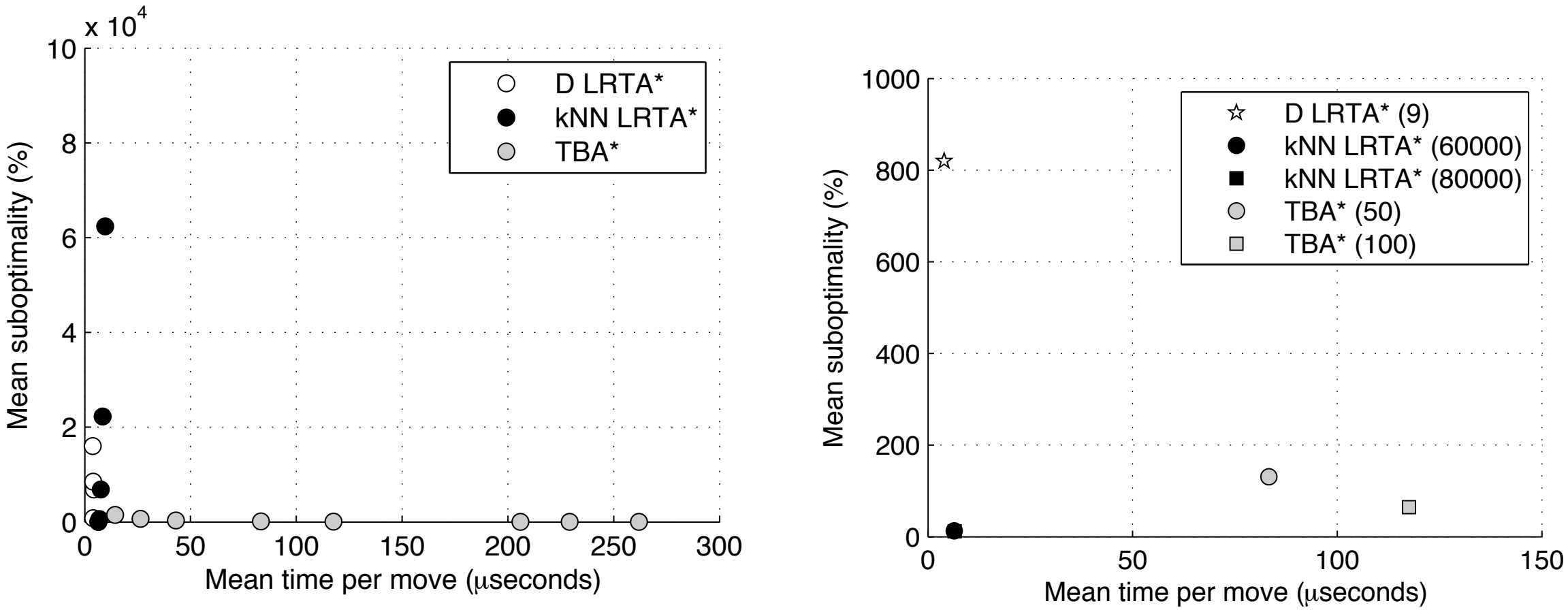

Figure 12: Suboptimality vs. time per move: all algorithms (**left**), high-performance region (**right**).

| **Algorithm** | **Mean time per move (microseconds)** | **Solution suboptimality (%)** |
|---|---|---|
| kNN LRTA*(10000) | 7.56 | 6851.62 |
| kNN LRTA*(40000) | 6.88 | 620.63 |
| kNN LRTA*(60000) | 6.40 | 12.77 |
| kNN LRTA*(80000) | 6.55 | 11.96 |
| D LRTA*(12) | 3.73 | 15999.23 |
| D LRTA*(11) | 3.93 | 8497.09 |
| D LRTA*(10) | 4.26 | 6831.74 |
| D LRTA*(9) | 3.94 | 819.72 |
| TBA*(5) | 14.31 | 1504.54 |
| TBA*(10) | 26.34 | 666.50 |
| TBA*(50) | 83.31 | 131.12 |
| TBA*(100) | 117.52 | 64.66 |
| A* | 208.03 | 0 |

Table 1: Suboptimality versus time per move.

it spends most of it time during the first move when it computes the entire path. Subsequent moves require a trivial computation. In the table, we define A*'s mean time per move as the total planning time for a problem divided by the number of moves in the path A* finds. We average this quantity over all problems. kNN LRTA* is about 30 times faster than A* per move.

Note that kNN LRTA*'s time per move decreases with larger databases. This is intuitive as with more database records there is a higher probability that an earlier record on the short list of records for whom the reachability checks are run will pass the checks (line 4 in Figure 8). Consequently, no further time-consuming reachability checks will be administered in the function **selectSubgoals**, saving time per move. These time savings, resulting from a larger database, outweigh the extra time spent traversing a correspondingly larger kd-tree to form the short list of most similar records. This fact indicates that the kd-tree approach scales well with the database size.

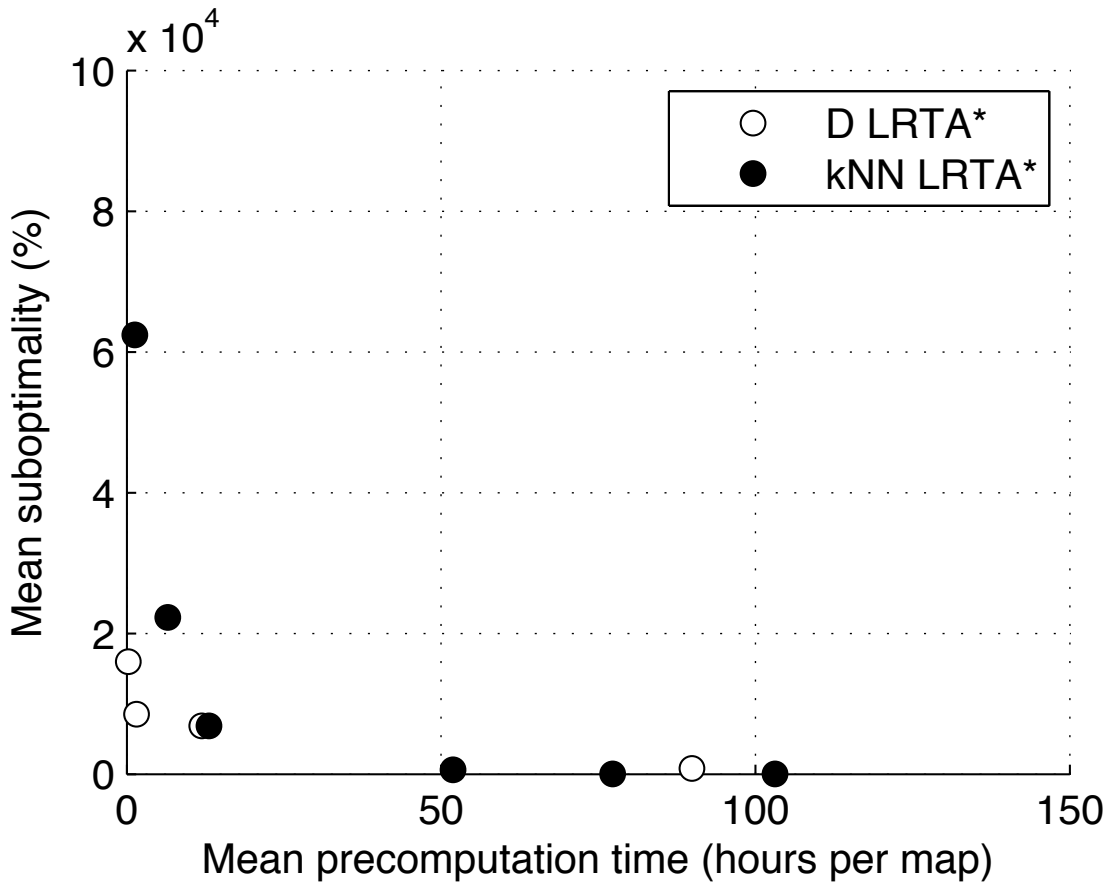

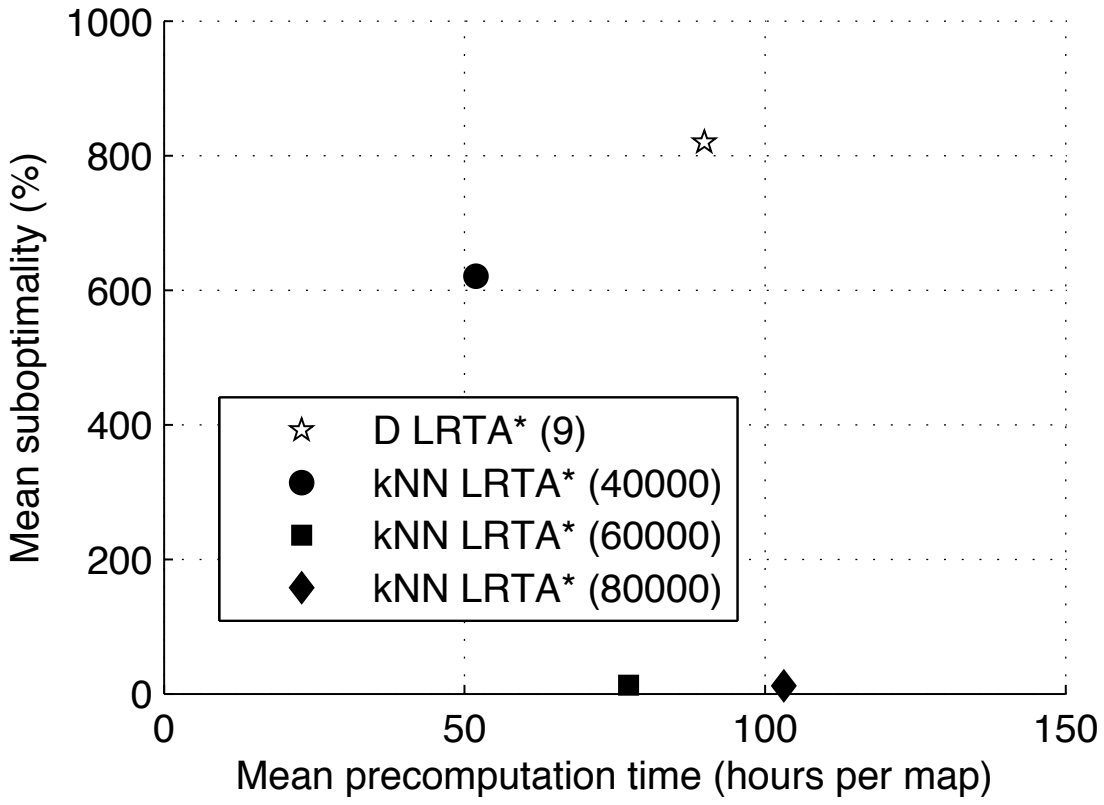

Figure 13: Suboptimality versus database pre-computation time per map. **Left:** all pre-computing algorithms. **Right:** a high-performance subplot.

| Algorithm | Pre-computation time per map (hours) | Solution suboptimality (%) |
|---|---|---|
| kNN LRTA*(10000) | 13.10 | 6851.62 |
| kNN LRTA*(40000) | 51.89 | 620.63 |
| kNN LRTA*(60000) | 77.30 | 12.77 |
| kNN LRTA*(80000) | 103.09 | 11.96 |
| D LRTA*(12) | 0.25 | 15999.23 |
| D LRTA*(11) | 1.57 | 8497.09 |
| D LRTA*(10) | 11.95 | 6831.74 |
| D LRTA*(9) | 89.88 | 819.72 |

Table 2: Suboptimality versus database pre-computation time.

### 8.4 Database Pre-computation Time

Suboptimality versus database pre-computation time is shown in Figure 13. The left subplot demonstrates all parametrizations of D LRTA* and kNN LRTA* while the right plot focuses on better performing configurations. Table 2 shows the individual values.

kNN LRTA* has three advantages over D LRTA*. First, kNN LRTA* with 40 and 60 thousand records easily dominates D LRTA*(9): it has better suboptimality while requiring less pre-computation time. kNN LRTA*(80000) is an overkill for these maps as it does not improve suboptimality by much (11.96% versus 12.77% achievable with 60000 records) while having the longest precomputation time.

Second, the database computation can be parallelized more easily in the case of kNN LRTA* as the individual records are completely independent of each other. This is not the case with D LRTA*. Additionally, D LRTA* requires building a map abstraction which is more complex to do in parallel.

Third, the number of records in the kNN LRTA* database can be controlled much more easily than in that of D LRTA*. Specifically, in D LRTA* one controls the level of abstraction. The number $S_a$ of abstract states at abstraction level $a$ is approximately $\frac{S}{\alpha^a}$ where $S$ is the number of original non-abstract states and $\alpha$ is a constant reduction factor (Bulitko et al., 2007). The number $N_a$ of records in D LRTA* database is $S_a(S_a - 1)$. Thus, the ratio between $N_a$ and $N_{a-1}$ is:

$$\frac{N_{a-1}}{N_a} = \frac{S_{a-1}(S_{a-1}-1)}{S_a(S_a-1)} = \frac{\frac{S}{\alpha^{a-1}}\left(\frac{S}{\alpha^{a-1}}-1\right)}{\frac{S}{\alpha^{a}}\left(\frac{S}{\alpha^{a}}-1\right)} = \frac{S-\alpha^{a-1}}{S-\alpha^{a}}\alpha^2 = \Omega(\alpha^2).$$

Thus by decreasing the level of abstraction by one, D LRTA* database size grows at least quadratically in $\alpha$. On our maps, clique abstraction has $\alpha$ of approximately 3 which means that there is nearly an order of magnitude in database size (and pre-computation time) when we go down by one level of abstraction. To illustrate, building a database for D LRTA*(8) is estimated to take over 800 CPU-hours. On the other hand, the number of records in the kNN LRTA* database is a user-specified parameter, affording a much greater control.

Of a particular interest is the pair kNN LRTA* with a database of 10000 and D LRTA* with abstraction level 10 as they perform closely in both measures. We discuss differences in their database sizes in the next section.

### 8.5 Database Size

Memory is at premium in video games, especially on consoles. TBA* space complexity comes from its open and closed list which it builds on-line. kNN LRTA* and D LRTA* expand only a single state (the agent's current state) and thus have the closed list of one state and the open list of at most eight states (as any grid cell in our maps has at most eight neighbors). However, these two algorithms consume memory as they store updated heuristic values. Additionally, they store their subgoal databases. In this section we focus on the database size. The next section will cover the total memory consumed on-line: open and closed lists as well as the updated heuristic values.

Each D LRTA* database record stores exactly three states. kNN LRTA* records have two or more states each and the number of records is fixed by the algorithm parameter. Additionally, kNN LRTA* stores start and end states of each record in a kd-tree. We define *relative database size* as the ratio of the total number of states stored in all records to the total number of map grid cells.

In addition to subgoal records, D LRTA* databases contain explicit region assignment for each state. Consequently, D LRTA* databases have a relative size of at least 1. This extra storage is a major weakness of D LRTA* in comparison to kNN LRTA*. To illustrate, as in our implementation we use 32 bits to index states, storing region assignment for each grid cell translates to an average of about 84 megabytes per map. Full results are found in Figure 14 and Table 3.

| Algorithm | Pre-computation time | Records | Relative size | Size (megabytes) |
|---|---|---|---|---|
| kNN LRTA*(10000) | 13.10 | 10000 | 0.00308 | 0.25 |
| kNN LRTA*(40000) | 51.89 | 40000 | 0.01234 | 1.00 |
| kNN LRTA*(60000) | 77.30 | 60000 | 0.01851 | 1.51 |
| kNN LRTA*(80000) | 103.09 | 80000 | 0.02468 | 2.01 |
| D LRTA*(12) | 0.25 | 251.5 | 1.00001 | 84.96 |
| D LRTA*(11) | 1.57 | 1896.5 | 1.00009 | 84.97 |
| D LRTA*(10) | 11.95 | 14872.0 | 1.00068 | 85.02 |
| D LRTA*(9) | 89.88 | 116048.5 | 1.00532 | 85.40 |

Table 3: Database statistics. All values are averages per map. Pre-computation time is in hours.

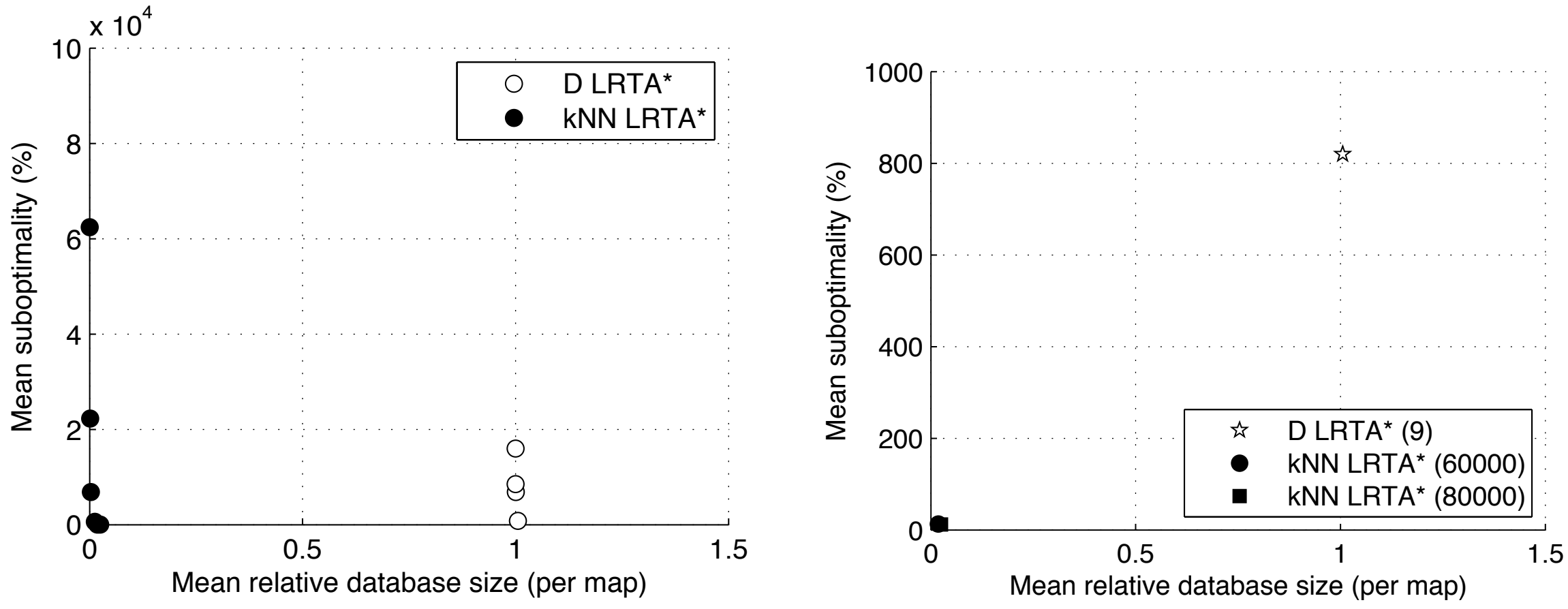

Figure 14: Suboptimality vs. database size: all algorithms (**left**), high-performance region (**right**).

Again, kNN LRTA* dominates D LRTA* by requiring much less memory and, at the same time, producing solutions of better quality. For instance, kNN LRTA*(60000) requires approximately 57 times less database memory than D LRTA*(9) while simultaneously producing solutions that are about eight times better.

Let us now re-visit the interesting case of kNN LRTA*(10000) and D LRTA*(10) which closely match each other with respect to database pre-computation time and solution suboptimality. As Table 3 reveals, kNN LRTA*(10000) uses approximately 340 times less memory than D LRTA*(10): about 256 kilobytes versus 85.02 megabytes.

Note that D LRTA*(10) averages approximately 49% more records per map than kNN LRTA*(10000) but requires approximately 9% less pre-computation time. This is because (i) D LRTA* averages fewer subgoals per record than kNN LRTA* and (ii) computing D LRTA* subgoals does not require reachability checks — a time-consuming process.

Also note that despite having 49% more records, D LRTA* affords only a 0.3% improvement in solution quality over kNN LRTA*. More generally, additional experiments have demonstrated that kNN LRTA* tends to outperform D LRTA* in solution quality given the same number of records. There are two factors in play here. First, kNN LRTA* records often contain several subgoals that are guaranteed to be reachable from each other without scrubbing. D LRTA* records offers no such guarantees and their only subgoal may be difficult to reach from the agent's start state when abstract regions become large and complex. On the upside, D LRTA* spaces its records in a systematic fashion — one record per each pair of regions — thereby providing a potentially better coverage than can be afforded by randomly selected starts and ends of kNN LRTA* database records. It appears that the former factor overcomes the latter, leading to kNN LRTA*'s better per-record suboptimality.

### 8.6 On-line Space Complexity

We will first analyze specifically the amount of memory allocated by the algorithms on-line. When an algorithm solves a particular problem, we record the maximum size of its open and closed lists as well as the total number of states whose heuristic values were updated. We count each updated heuristic value as one state in terms of storage required.[5] Adding these three measures together, we

5. Multiple heuristic updates in the same state do not increase the amount of storage.

| Algorithm | Strictly on-line memory (Kbytes) | Solution suboptimality (%) |
|---|---|---|
| kNN LRTA*(10000) | 8.62 | 6851.62 |
| kNN LRTA*(40000) | 5.04 | 620.63 |
| kNN LRTA*(60000) | 4.23 | 12.77 |
| kNN LRTA*(80000) | 4.22 | 11.96 |
| D LRTA*(12) | 18.76 | 15999.23 |
| D LRTA*(11) | 11.09 | 8497.09 |
| D LRTA*(10) | 8.24 | 6831.74 |
| D LRTA*(9) | 3.04 | 819.72 |
| TBA*(5) | 1353.94 | 1504.54 |
| TBA*(10) | 1353.94 | 666.50 |
| TBA*(50) | 1353.94 | 83.31 |
| TBA*(100) | 1353.94 | 64.66 |
| A* | 1353.94 | 0 |

Table 4: Strictly on-line memory versus solution suboptimality.

record the amount of *strictly on-line memory* per problem. Averaging the strictly on-line memory over all problems, we list the results in Table 4.

kNN LRTA* dominates all D LRTA* points except D LRTA*(9) which has the lowest mean strictly on-line memory of 3.04 Kbytes per problem. TBA*, being effectively a time-sliced A*, does not update heuristic values at all. However, its open and closed lists contribute to the highest memory consumption at 1353.94 Kbytes. This is intuitive as TBA* does not use subgoals and therefore must "fill in" potentially large heuristic depressions with its open and closed lists. Also, notice that the total size of these lists does not change with the cutoff as state expansions are independent of agent's moves in TBA*. A* has identical memory consumption as it expands states in the same way as TBA*. Again, kNN LRTA* dominates TBA* for all cutoff values, using less memory and producing better solutions.

Strictly on-line memory gives an insight into the algorithms but does not present a complete picture. Specifically, D LRTA* and kNN LRTA* must load their databases into their on-line memory. Thus we define the *cumulative on-line memory* as the strictly on-line memory plus the size of the database loaded. The values are found in Figure 15 and Table 5.

Several observations are due. First, TBA* is no longer dominated due to its low memory consumption. Second, D LRTA* is in its own league due to explicitly labelling every state with a corresponding region as well as computing subgoals for all pairs of regions. Third, D LRTA* has a sweet spot in its memory consumption which corresponds to the abstraction level of 11. A higher level of abstraction reduces the database size but not enough to compensate for more updated heuristic values. Lower abstraction levels reduce the amount of learning but not enough to compensate for large number of subgoals in the database.

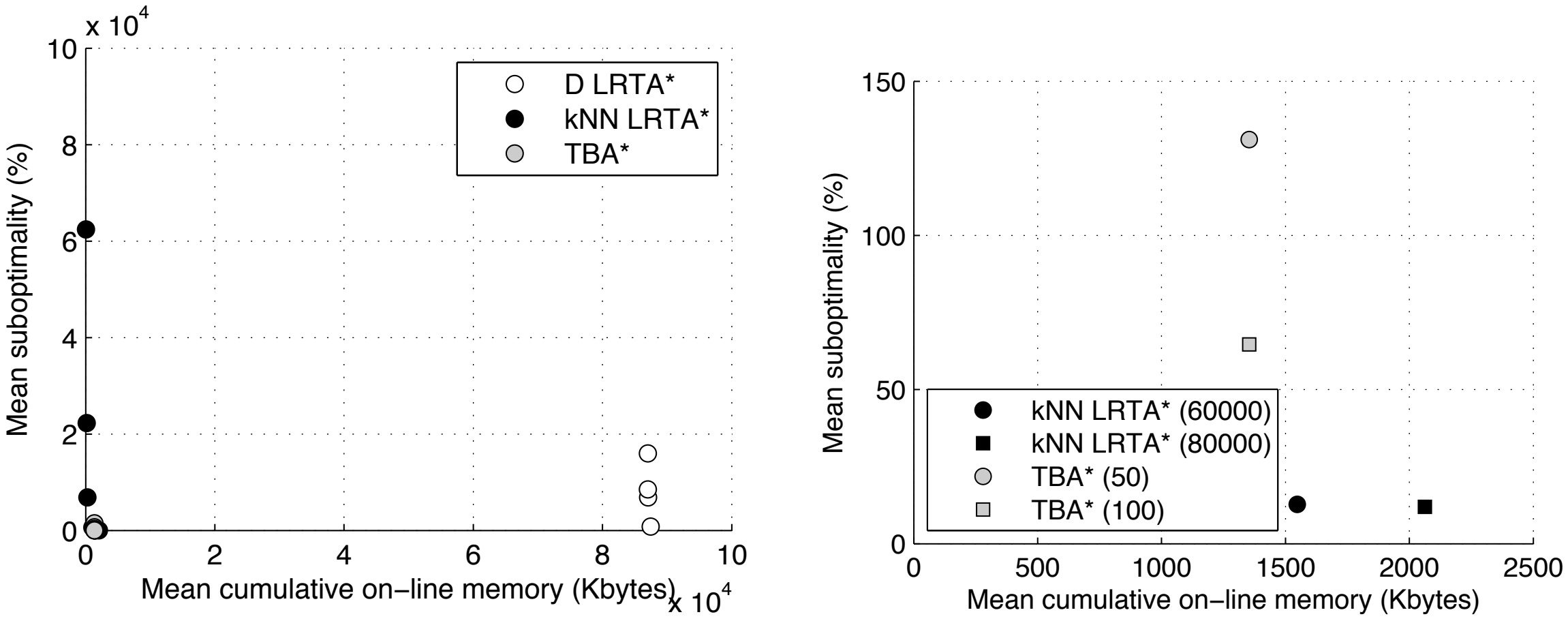

Figure 15: Suboptimality versus cumulative on-line memory. **Left:** all algorithms. **Right:** a high-performance subplot.

| Algorithm | Cumulative on-line memory (Kbytes) | Solution suboptimality (%) |
|---|---|---|
| kNN LRTA*(10000) | 265.65 | 6851.62 |
| kNN LRTA*(40000) | 1034.08 | 620.63 |
| kNN LRTA*(60000) | 1547.85 | 12.77 |
| kNN LRTA*(80000) | 2062.20 | 11.96 |
| D LRTA*(12) | 87019.74 | 15999.23 |
| D LRTA*(11) | 87018.50 | 8497.09 |
| D LRTA*(10) | 87066.34 | 6831.74 |
| D LRTA*(9) | 87456.35 | 819.72 |
| TBA*(5) | 1353.94 | 1504.54 |
| TBA*(10) | 1353.94 | 666.50 |
| TBA*(50) | 1353.94 | 83.31 |
| TBA*(100) | 1353.94 | 64.66 |
| A* | 1353.94 | 0 |

Table 5: Solution suboptimality versus cumulative on-line memory.

### 8.7 Simultaneous Pathfinding by Multiple Agents

If only a single agent is pathfinding at a time, the analysis above holds and TBA* is the most memory efficient choice. However, in most video games, anywhere from half a dozen to a thousand agents (e.g., Gas Powered Games, 2007) can be pathfinding simultaneously on the same map.

Such a scenario favors kNN LRTA* and D LRTA* whose subgoal databases are map-specific but independent of start and goal states. Consequently, multiple agents running D LRTA* and kNN LRTA* can all share the same subgoal database.[6] In contrast, all memory consumed by TBA* is specific to a given agent and cannot be shared with other agents operating on the same map.

6. Note that such multiple agents cannot, generally speaking, share their heuristic $h$ as it is computed and updated with respect to different goals.

Hence the total amount of cumulative on-line memory for $K$ agents operating simultaneously equals the amount of the database memory plus $K$ times the amount of strictly on-line memory. A *break-even point* for an algorithm $A$ with respect to an algorithm $B$ is then defined as the minimal number of agents using $A$ that collectively consume less memory than the same number of agents using $B$. Table 6 lists the break-even points for D LRTA* and kNN LRTA* with respect to TBA*.

| **Algorithm** | **Break-even point with respect to TBA* (number of agents)** |
|---|---|
| kNN LRTA*(10000) | 1 |
| kNN LRTA*(40000) | 1 |
| kNN LRTA*(60000) | 2 |
| kNN LRTA*(80000) | 2 |
| D LRTA*(9) | 65 |
| D LRTA*(10) | 65 |
| D LRTA*(11) | 66 |
| D LRTA*(12) | 66 |

Table 6: Break-even points for kNN LRTA* and D LRTA* with respect to TBA*.

kNN LRTA* with ten and forty thousand record databases requires less cumulative on-line memory than TBA* and hence the break-even point is just one agent. For sixty and eighty thousand records, two kNN LRTA* agents take less total cumulative on-line memory than two TBA* agents. It takes 65 to 66 simultaneously pathfinding agents to amortize the large D LRTA* databases and gain a memory advantage over TBA*.

## 9. Discussion

This is the first time the high-performance real-time search algorithms TBA*, D LRTA* and kNN LRTA* were evaluated on contemporarily sized maps. The results, presented in detail in the previous section, are summarized for representative algorithms in Table 7.

| **Dimension** | **kNN LRTA* versus other algorithms** |
|---|---|
| Suboptimality | kNN LRTA* is 8.16 times better than D LRTA* and 1.46 times better than TBA* |
| Time per move | kNN LRTA* is 18.36 times better than TBA* and 62% worse than D LRTA* |
| Cumulative memory | kNN LRTA* is 57 times better than D LRTA* and 13% times worse than TBA* |
| Break-even point | kNN LRTA* takes less memory than TBA* with two or more agents |
| Pre-computation time | kNN LRTA* is 14% better than D LRTA* |

Table 7: Comparisons of kNN LRTA*(60000) to D LRTA*(9) and TBA*(100).

kNN LRTA* achieves the best suboptimality of the three algorithms. kNN LRTA* is substantially faster per move than TBA* and is on par with D LRTA*. In terms of cumulative on-line memory, kNN LRTA* outperforms D LRTA* by two orders of magnitude and is only 13% worse than TBA*. Furthermore, with two or more simultaneously planning agents, kNN LRTA* takes less memory than TBA*. In contrast, it takes 65 or more D LRTA* agents to amortize its database and gain a memory advantage over TBA*. Off-line, kNN LRTA* outperforms D LRTA* by achieving an order of magnitude better solutions with a database two orders of magnitude smaller in size and slightly faster to compute.

While the results of comparisons to TBA* were expected as TBA* does not benefit from pre-computation, the comparison between kNN LRTA* and D LRTA* unearthed unexpected results. Specifically, subgoal databases of kNN LRTA* are a more effective use of pre-computation time and memory than those of D LRTA*. The lower memory consumption with kNN LRTA* databases is achieved by not having to store explicit region membership. Better pre-computation times come from not having to compute shortest paths for all pairs of abstract times. Finally, kNN LRTA* is better than D LRTA* on a per record basis. This is because compressing an entire optimal path into a series of subgoals reachable from each other via hill-climbing guarantees that once a single subgoal is reached, the underlying LRTA* agent will reach its global goal without scrubbing. In contrast, D LRTA* subgoals can be difficult to reach from the agent's current position. And even when reached, the difficulties can recur with the subsequent subgoals.

In terms of applications, kNN LRTA* can be the algorithm of choice to use in video game pathfinding. For instance, kNN LRTA*(60000) is over 30 times faster per average move than commonly used A* and produces solutions which are less than 15% suboptimal. This performance comes at the cost of about 77 hours of pre-computation time per map which can be easily reduced to under 10 hours on a modern eight-core workstation. This is negligible comparing to the amount of time a game company spends hand-crafting a single map.

## 10. Current Shortcomings and Future Work

Despite outperforming existing state-of-the-art real-time search algorithms on our problems overall, kNN LRTA* has several shortcomings. First, the database records are generated off randomly selected start and end states. This means that the coverage of the space is not necessarily even: some small, but difficult to reach, regions of the space may never get a suitable record while some easy to reach regions may get multiple redundant records covering it.

Increasing database efficiency would allow a smaller database to afford an equal coverage and hence an equal on-line performance. This will in turn reduce the pre-computation time of kNN LRTA* database which can presently reach over 100 hours per map. While the computation can be sped up at a nearly linear scale by using multi-core processors and the time is affordable on the game company side, most players would want their home-made game maps to be processed in a matter of seconds or minutes.

Making the subgoal coverage more uniform can be accomplished via forgoing random start and end selection in favor of space partitioning. However, unlike D LRTA*'s abstract regions built via repeated applications of the clique abstraction, such partitions should have all their states reachable from each other by LRTA* without scrubbing. Then start and end states for the database records can be selected within these partitions. To reduce the amount of pre-computation one can then compute subgoals by compressing optimal paths only between neighboring regions (as opposed to all distinct abstract regions as D LRTA* does). Note that unlike D LRTA*, the partitioning is necessarily only off-line and no explicit region assignment is stored for every state. As a result, the on-line memory consumption will be comparable or better than the existing kNN LRTA*.

A philosophically oriented project would be to develop a self-aware agent. Specifically, such an agent would analyze the performance of its core algorithm (e.g., LRTA*) and decide on the appropriate partitioning scheme. A similar meta-level control has been previously attempted for dynamic selection of lookahead depth in real-time search (e.g., Russell & Wefald, 1991).

## 11. Beyond Grid Pathfinding

We presented and evaluated kNN LRTA* for grid-based pathfinding. Formally, the algorithm, with the exception of its kd-tree module, is applicable to arbitrary weighted graphs that satisfy the constraints at the beginning of Section 2. In principle, it should be applicable to general planning using the ideas from search-based planners ASP (Bonet, Loerincs, & Geffner, 1997), the HSP-family (Bonet & Geffner, 2001), FF (Hoffmann, 2000), SHERPA (Koenig, Furcy, & Bauer, 2002) and LDFS (Bonet & Geffner, 2006).

As we described earlier in the paper, using the kd-tree index requires a certain correspondence between coordinate similarity and heuristic distance. Extending kd-trees or developing appropriate new index structures for an arbitrary graph is an open research question. An interim solution is to apply kNN LRTA* to arbitrary search problems without its kd-tree module. Doing so will, however, slow down the on-line part because similarity must be computed between agent's current situation and every single record in the database. On the positive side, not computing kd-trees will speed up the off-line part of kNN LRTA*.

Finally, while kNN LRTA* is theoretically applicable to arbitrary search problems, it is not clear how well it will perform there with respect to its competitors D LRTA* and TBA*. Such an investigation is left for future work.

## 12. Conclusions

In this paper we considered the problem of real-time heuristic search whose planning time per move does not depend on the number of states. We proposed a new mechanism for selecting subgoals automatically. The resulting algorithm was shown to be theoretically complete and, on large video game maps, substantially outperformed the previous state-of-the-art algorithms D LRTA* and TBA* along several important performance measures.

## Acknowledgments

This research was supported by grants from the National Science and Engineering Research Council of Canada (NSERC); Icelandic Centre for Research (RANNÍS); and by a Marie Curie Fellowship of the European Community programme *Structuring the ERA* under contract number MIRG-CT-2005-017284. We appreciate help of Josh Sterling, Stephen Hladky and Daniel Huntley.

## References

Anwar, M. A., & Yoshida, T. (2001). Integrating OO road network database, cases and knowledge for route finding. In *ACM Symposium on Applied Computing (SAC)*, pp. 215–219. ACM.

Barto, A. G., Bradtke, S. J., & Singh, S. P. (1995). Learning to act using real-time dynamic programming. *Artificial Intelligence*, *72*(1), 81–138.

BioWare Corp. (1998). Baldur's Gate., Published by Interplay, http://www.bioware.com/bgate/, November 30, 1998.

Björnsson, Y., Bulitko, V., & Sturtevant, N. (2009). TBA*: Time-bounded A*. In *Proceedings of the International Joint Conference on Artificial Intelligence (IJCAI)*, pp. 431 – 436, Pasadena, California. AAAI Press.

Björnsson, Y., & Halldórsson, K. (2006). Improved heuristics for optimal path-finding on game maps. In Laird, J. E., & Schaeffer, J. (Eds.), *Proceedings of the Second Artificial Intelligence and Interactive Digital Entertainment Conference (AIIDE), June 20-23, 2006, Marina del Rey, California*, pp. 9–14. The AAAI Press.

Blizzard Entertainment (2002). Warcraft III: Reign of chaos., Published by Blizzard Entertainment, http://www.blizzard.com/war3, July 3, 2002.

Bonet, B., & Geffner, H. (2001). Planning as heuristic search. *Artificial Intelligence*, *129*(1–2), 5–33.

Bonet, B., & Geffner, H. (2006). Learning depth-first search: A unified approach to heuristic search in deterministic and non-deterministic settings, and its application to MDPs. In *Proceedings of the International Conference on Automated Planning and Scheduling (ICAPS)*, pp. 142–151, Cumbria, UK.

Bonet, B., Loerincs, G., & Geffner, H. (1997). A fast and robust action selection mechanism for planning. In *Proceedings of the National Conference on Artificial Intelligence (AAAI)*, pp. 714–719, Providence, Rhode Island. AAAI Press / MIT Press.

Branting, K., & Aha, D. W. (1995). Stratified case-based reasoning: Reusing hierarchical problem solving episodes. In *Proceedings of the International Joint Conference on Artificial Intelligence (IJCAI)*, pp. 384–390.

Bulitko, V. (2004). Learning for adaptive real-time search. Tech. rep. http://arxiv.org/abs/cs.AI/0407016, Computer Science Research Repository (CoRR).

Bulitko, V., & Björnsson, Y. (2009). kNN LRTA*: Simple subgoaling for real-time search. In *Proceedings of Artificial Intelligence and Interactive Digital Entertainment (AIIDE)*, pp. 2–7, Stanford, California. AAAI Press.

Bulitko, V., Björnsson, Y., Luštrek, M., Schaeffer, J., & Sigmundarson, S. (2007). Dynamic Control in Path-Planning with Real-Time Heuristic Search. In *Proceedings of the International Conference on Automated Planning and Scheduling (ICAPS)*, pp. 49–56, Providence, RI.

Bulitko, V., & Lee, G. (2006). Learning in real time search: A unifying framework. *Journal of Artificial Intelligence Research (JAIR)*, *25*, 119–157.

Bulitko, V., Luštrek, M., Schaeffer, J., Björnsson, Y., & Sigmundarson, S. (2008). Dynamic control in real-time heuristic search. *Journal of Artificial Intelligence Research (JAIR)*, *32*, 419 – 452.

Bulitko, V., Sturtevant, N., Lu, J., & Yau, T. (2007). Graph abstraction in real-time heuristic search. *Journal of Artificial Intelligence Research (JAIR)*, *30*, 51–100.

Carbonell, J. G., Knoblock, C., & Minton, S. (1990). Prodigy: An integrated architecture for planning and learning. In Lehn, K. V. (Ed.), *Architectures for Intelligence*. Lawrence Erlbaum Associates.

Cazenave, T. (2006). Optimizations of data structures, heuristics and algorithms for path-finding on maps. In Louis, S. J., & Kendall, G. (Eds.), *Proceedings of the 2006 IEEE Symposium* *on Computational Intelligence and Games (CIG06), University of Nevada, Reno, campus in Reno/Lake Tahoe, 22-24 May, 2006*, pp. 27–33. IEEE.

Culberson, J., & Schaeffer, J. (1998). Pattern Databases. *Computational Intelligence*, *14*(3), 318–334.

Furcy, D., & Koenig, S. (2000). Speeding up the convergence of real-time search. In *Proceedings of the National Conference on Artificial Intelligence (AAAI)*, pp. 891–897.

Gas Powered Games (2007). Supreme Commander., Published by THQ, http://www.supremecommander.com/, February 20, 2007.

Geisberger, R., Sanders, P., Schultes, D., & Delling, D. (2008). Contraction hierarchies: Faster and simpler hierarchical routing in road networks. In McGeoch, C. C. (Ed.), *WEA*, Vol. 5038 of *Lecture Notes in Computer Science*, pp. 319–333. Springer.

Haigh, K., & Veloso, M. (1993). Combining search and analogical reasoning in path planning from road maps. In *Proceedings of the AAAI-93 Workshop on Case-Based Reasoning*, pp. 79–85, Washington, DC. AAAI. AAAI Press technical report WS-93-01.

Hart, P., Nilsson, N., & Raphael, B. (1968). A formal basis for the heuristic determination of minimum cost paths. *IEEE Transactions on Systems Science and Cybernetics*, *4*(2), 100–107.

Hernández, C., & Meseguer, P. (2005a). Improving convergence of LRTA*(k). In *Proceedings of the International Joint Conference on Artificial Intelligence (IJCAI), Workshop on Planning and Learning in A Priori Unknown or Dynamic Domains*, pp. 69–75, Edinburgh, UK.

Hernández, C., & Meseguer, P. (2005b). LRTA*(k). In *Proceedings of the International Joint Conference on Artificial Intelligence (IJCAI)*, pp. 1238–1243, Edinburgh, UK.

Hodal, J., & Dvorak, J. (2008). Using case-based reasoning for mobile robot path planning. *Engineering Mechanics*, *15*, 181–191.

Hoffmann, J. (2000). A heuristic for domain independent planning and its use in an enforced hill-climbing algorithm. In *Proceedings of the 12th International Symposium on Methodologies for Intelligent Systems (ISMIS)*, pp. 216–227.

Ishida, T. (1992). Moving target search with intelligence. In *National Conference on Artificial Intelligence (AAAI)*, pp. 525–532.

Koenig, S. (2004). A comparison of fast search methods for real-time situated agents. In *Proceedings of Int. Joint Conf. on Autonomous Agents and Multiagent Systems*, pp. 864 – 871.

Koenig, S., Furcy, D., & Bauer, C. (2002). Heuristic search-based replanning. In *Proceedings of the Int. Conference on Artificial Intelligence Planning and Scheduling*, pp. 294–301.

Koenig, S., & Likhachev, M. (2006). Real-time adaptive A*. In *Proceedings of the International Joint Conference on Autonomous Agents and Multiagent Systems (AAMAS)*, pp. 281–288.

Korf, R. (1985). Depth-first iterative deepening: An optimal admissible tree search. *Artificial Intelligence*, *27*(3), 97–109.

Korf, R. (1990). Real-time heuristic search. *Artificial Intelligence*, *42*(2–3), 189–211.

Likhachev, M., Ferguson, D. I., Gordon, G. J., Stentz, A., & Thrun, S. (2005). Anytime dynamic A*: An anytime, replanning algorithm. In *ICAPS*, pp. 262–271.

Likhachev, M., Gordon, G. J., & Thrun, S. (2004). ARA*: Anytime A* with provable bounds on sub-optimality. In Thrun, S., Saul, L., & Schölkopf, B. (Eds.), *Advances in Neural Information Processing Systems 16*. MIT Press, Cambridge, MA.

Luštrek, M., & Bulitko, V. (2006). Lookahead pathology in real-time path-finding. In *Proceedings of the National Conference on Artificial Intelligence (AAAI), Workshop on Learning For Search*, pp. 108–114, Boston, Massachusetts.

Moore, A. (1991). *Efficient Memory-based Learning for Robot Control*. Ph.D. thesis, University of Cambridge.

Nebel, B., & Koehler, J. (1995). Plan reuse versus plan generation: A theoretical and empirical analysis. *Artificial Intelligence*, *76*, 427–454.

Rayner, D. C., Davison, K., Bulitko, V., Anderson, K., & Lu, J. (2007). Real-time heuristic search with a priority queue. In *Proceedings of the International Joint Conference on Artificial Intelligence (IJCAI)*, pp. 2372–2377, Hyderabad, India.

Russell, S., & Wefald, E. (1991). *Do the right thing: Studies in limited rationality*. MIT Press.

Shimbo, M., & Ishida, T. (2003). Controlling the learning process of real-time heuristic search. *Artificial Intelligence*, *146*(1), 1–41.

Shue, L.-Y., Li, S.-T., & Zamani, R. (2001). An intelligent heuristic algorithm for project scheduling problems. In *Proceedings of the 32nd Annual Meeting of the Decision Sciences Institute*, San Francisco.

Shue, L.-Y., & Zamani, R. (1993). An admissible heuristic search algorithm. In *Proceedings of the 7th International Symposium on Methodologies for Intelligent Systems (ISMIS-93)*, Vol. 689 of *LNAI*, pp. 69–75.

Sigmundarson, S., & Björnsson, Y. (2006). Value Back-Propagation vs. Backtracking in Real-Time Search. In *Proceedings of the National Conference on Artificial Intelligence (AAAI), Workshop on Learning For Search*, pp. 136–141, Boston, Massachusetts, USA.

Stenz, A. (1995). The focussed D* algorithm for real-time replanning. In *Proceedings of the International Joint Conference on Artificial Intelligence (IJCAI)*, pp. 1652–1659.

Sturtevant, N. (2007). Memory-efficient abstractions for pathfinding. In *Proceedings of the third conference on Artificial Intelligence and Interactive Digital Entertainment*, pp. 31–36, Stanford, California.

Sturtevant, N., & Buro, M. (2005). Partial pathfinding using map abstraction and refinement. In *Proceedings of the National Conference on Artificial Intelligence (AAAI)*, pp. 1392–1397, Pittsburgh, Pennsylvania.

Sturtevant, N. R., Felner, A., Barrer, M., Schaeffer, J., & Burch, N. (2009). Memory-based heuristics for explicit state spaces. In Boutilier, C. (Ed.), *IJCAI 2009, Proceedings of the 21st International Joint Conference on Artificial Intelligence, Pasadena, California, USA, July 11-17, 2009*, pp. 609–614.

Valve Corporation (2004). Counter-Strike: Source., Published by Valve Corporation, http://store.steampowered.com/app/240/, October 7, 2004.

Weng, M., Wei, X., Qu, R., & Cai, Z. (2009). A path planning algorithm based on typical case reasoning. *Geo-spatial Information Science*, *12*, 66–71.